\documentclass{article}

\usepackage{microtype}
\usepackage{graphicx}
\usepackage{subfigure}
\usepackage{booktabs} 

\usepackage{hyperref}

\usepackage[accepted]{icml2025}

\usepackage{amsmath}
\usepackage{amssymb}
\usepackage{mathtools}
\usepackage{amsthm}

\usepackage[capitalize,noabbrev]{cleveref}

\theoremstyle{plain}

\theoremstyle{definition}

\theoremstyle{remark}

\usepackage[textsize=tiny]{todonotes}

\usepackage{array}
\usepackage{arydshln}  
\usepackage{algorithm}
\usepackage{algorithmic}
\usepackage{arydshln}
\usepackage{amsmath}
\usepackage{amsthm} 
\usepackage{booktabs}
\usepackage{comment}
\usepackage{caption}
\usepackage{float}
\usepackage{graphicx}
\usepackage{hyperref}
\usepackage{makecell}
\usepackage{multirow} 
\usepackage{subfigure}
\usepackage{url}
\usepackage{wrapfig}
\usepackage{adjustbox}

\usepackage{tikz}
\usetikzlibrary{shapes.geometric, arrows}
\usepackage{amsmath}
\usepackage{amssymb}
\usepackage{soul}
\usepackage{mdframed}
\definecolor{lightgray}{gray}{0.9}
\usepackage{bm} 
\newcommand{\bep}{\bm{\epsilon}}
\newcommand{\bsx}{\bm{x}}
\newcommand{\bsm}{\bm{m}}
\newcommand{\bsl}{\bm{l}}
\newcommand{\bsP}{\bm{P}}

\usepackage{setspace} 

\let\oldtextcolor\textcolor
\renewcommand{\textcolor}[2]{#2}
\newcommand{\keepcolor}[2]{\oldtextcolor{#1}{#2}} 

\hypersetup{
    colorlinks=true,
    linkcolor=red,  
    urlcolor=magenta,
    filecolor=magenta,
    citecolor=blue,  
    pdftitle={Overleaf Example},
    pdfpagemode=FullScreen,
}

\def\bcP{{\boldsymbol{\mathcal{P}}}}
\def\bsm{{\boldsymbol{m}}}
\def\bsx{{\boldsymbol{x}}}

\def\bsP{{\boldsymbol{P}}}
\def\bsp{{\boldsymbol{p}}}

\def\bep{{\boldsymbol{\epsilon}}}

\def \xc #1{\textcolor{black}{#1}}

\icmltitlerunning{Bootstrapping Text to Control Time-Series Generation}

\begin{document}
\newcommand{\frameworkname}{\textsc{\sffamily\selectfont Bridge}}

\twocolumn[
\icmltitle{BRIDGE: Bootstrapping Text to Control Time-Series Generation \\
           via Multi-Agent Iterative Optimization and Diffusion Modeling}



\icmlsetsymbol{equal}{*}
\icmlsetsymbol{intern}{†}

\begin{icmlauthorlist}
\icmlauthor{Hao Li}{equal,microsoft,manchester} 
\icmlauthor{Yu-Hao Huang}{equal,microsoft,nanjing}
\icmlauthor{Chang Xu}{microsoft}
\icmlauthor{Viktor Schlegel}{manchester,ic} \\
\icmlauthor{Renhe Jiang}{tokyo} 
\icmlauthor{Riza Batista-Navarro}{manchester}
\icmlauthor{Goran Nenadic}{manchester}
\icmlauthor{Jiang Bian}{microsoft}
\end{icmlauthorlist}

\icmlaffiliation{microsoft}{Microsoft Research}
\icmlaffiliation{manchester}{The University of Manchester, UK}
\icmlaffiliation{nanjing}{Nanjing University, China}
\icmlaffiliation{ic}{Imperial College
London, Imperial Global Singapore, Singapore}
\icmlaffiliation{tokyo}{The University of Tokyo, Japan}

\icmlcorrespondingauthor{Chang Xu}{chanx@microsoft.com}

\vskip 0.3in
]



\printAffiliationsAndNotice{\icmlEqualContribution} 

\begin{abstract}
Time-series Generation (TSG) is \xc{a prominent research area 
with broad applications in simulations, data augmentation, and counterfactual analysis.}
While existing methods have shown promise in unconditional single-domain TSG, real-world applications demand
\xc{for cross-domain approaches capable of controlled}
generation tailored to domain-specific constraints and instance-level requirements. 
In this paper, we argue that text can provide semantic insights, \xc{domain information and instance-specific temporal patterns}, to \xc{guide and improve TSG}. 
We introduce ``Text-Controlled TSG'', a task focused on generating realistic time series by incorporating textual descriptions.
To address data scarcity in this setting, we propose a novel LLM-based Multi-Agent framework that synthesizes diverse, realistic text-to-TS datasets.
\xc{Furthermore, we introduce \frameworkname{}, a hybrid text-controlled TSG framework that integrates semantic prototypes with text description for supporting domain-level guidance.
This approach achieves state-of-the-art generation fidelity on 11 of 12 datasets, and improves controllability by up to 12\% on MSE and 6\% MAE compared to no text input generation,
highlighting its potential for generating tailored time-series data. Our code is available at: \href{https://github.com/microsoft/TimeCraft/tree/main/BRIDGE}{\texttt{Microsoft/TimeCraft}}}\footnote{Work done during Hao Li and Yu-Hao Huang's research internship at Microsoft Research.}.
\end{abstract}
\vspace{-0.3in}
\section{Introduction}

High-quality Time Series Generation (TSG) is an important task in various domains, including finance \citep{DBLP:journals/asc/SezerGO20}, healthcare \citep{DBLP:conf/bionlp/LiWSBNKZB0N23} and environmental science \citep{hasnain2022time}. 
For example, realistic synthetic medical electrocardiogram (ECG) patterns can be used to train medical residents \citep{pohl2025generating, DBLP:journals/bspc/HongC23}, while simulating regional electricity usage can be used to stress test the power grid \citep{westgaard2021performing}. 
Although some remarkable works \citep{DBLP:journals/nn/HuangD23, DBLP:journals/corr/abs-2403-03698} have been done for TSG, showing promising results in generating realistic and coherent time series (TS), most of them focus on the basic setting---unconditional single domain generation. 
However, in real application scenarios, there are specific constraints or requirements for the generated TS to be met, such as specifying domain-specific characteristics, incorporating prior knowledge \citep{DBLP:conf/iclr/YuanQ24}, or satisfying operational constraints \citep{DBLP:conf/nips/ColettaGBV23}. 
For instance, it may be necessary to generate ECG patterns that respect individual patient profiles or capture specific disease conditions \citep{DBLP:conf/clef/SchlegelLW0NKBZ23}. 
Therefore, the current status in TSG, while demonstrating strong foundational performance, leaves a significant gap for addressing more complex, constrained generation tasks that are crucial for real-world  applications.

Some prior work on cross-domain TSG has explored  various ways to meet specific generation needs, with most focusing on leveraging domain information to control the generation process. 
Some approaches rely on explicit domain labels during training \citep{DBLP:journals/corr/abs-2408-12991, kollovieh2024predict}, treating the task as a conditional generation problem. 
This allows users to specify the domain during inference. 
However, this method is limited as it struggles with \emph{unseen} domains and becomes inefficient when the number of domains is large. 
Other methods incorporate specific information through natural language \citep{DBLP:conf/infocom/ZhouJHX0Y24, liu2024unitime}, 
but they operate at the domain level, thus failing to provide detailed fine-grained and instance-specific control, which is essential for more accurate and tailored TSG, highlighting a significant gap in the field.

In this work, we investigate the challenging yet practical research problem of achieving \emph{instance-level} controlled TSG capable of \emph{generalising to unseen domains}.
Inspired by the recent success of controlled content generation in images \citep{DBLP:journals/corr/abs-2310-02239} and videos \citep{DBLP:journals/corr/abs-2402-17177}, where texts are used as a source of control which facilitates capturing complex patterns and semantic relationships, 
we argue that \textit{using text to provide semantic insights—such as \xc{domain information}
and instance-specific temporal patterns—could enhance and guide TSG}.
However, using text for controlled TSG presents two key challenges that need to be addressed in order to fully leverage its potential.

\emph{(i)} \textbf{Limited availability of high-quality text-TS pairs}: To train a TSG model that can be controlled by text, we require paired data, where each time series is associated with a detailed text description. However, most available text data only provides high-level domain descriptions, lacking granular, instance-specific information such as trends, fluctuations, or the behavior of individual data points \cite{DBLP:conf/kdd/LiangWNJ0SPW24}. The first challenge, therefore, is determining \textit{what specific text information is useful for controlling generation} and \textit{how to obtain such detailed text}. 
We explored rule-based methods to generate individual-level text descriptions 
\citep{DBLP:journals/corr/abs-2207-05194}, but it did not lead to significant performance improvements (as shown in Appendix~\ref{LLM_directly}), suggesting that a more sophisticated approach is needed. 

\emph{(ii)} \xc{\textbf{Bridging discrepancy between text and time-series data for controlled TSG}:}

\xc{Text and time-series (TS) data exhibit significant differences in both modality and granularity. Text conveys information via a fixed vocabulary of discrete tokens, while time series data is continuous, which leads to inherent mismatches.
This disparity may render text too coarse to fully capture domain-specific patterns and characteristics, posing challenges in achieving precise domain-level control due to its incomplete or oversimplified representations. 
Meanwhile, text can provide detailed, instance-specific descriptions that are crucial for nuanced control, requiring careful and dedicated design to align text with TS features effectively.}

\textbf{To address the first challenge}, we propose a role-based collaborative multi-agent framework designed to generate high-quality datasets for text controlled TSG. We argue that the process of automatically identifying textual descriptions for TS parallels prompt optimization for large language models (LLMs), where variations in prompt design significantly affect performance \citep{DBLP:conf/spcom/TASSAGKK24}. Inspired by the success of prior work~\citep{DBLP:conf/iclr/ZhouMHPPCB23, DBLP:conf/cec/LiuCQ0O24, DBLP:conf/iclr/Guo0GLS0L0Y24}, our framework consists of three key components: \textit{Text Template Generation}, \textit{Automatic Evaluation}, and \textit{Feedback-driven Refinement}. This process ensures continuous improvement through feedback and synthesis, achieving a holistic optimization of the generated text tailored to the TS at hand. Experimental evaluations reveal the effectiveness of the proposed framework, achieving at least a 15\% performance improvement in MAE compare to the unrefined text, while producing outputs that are notably more comprehensive than those generated through traditional text-generation methods. 

\textbf{To address the second challenge}, we adopt a hybrid text-enhanced time-series generation strategy. 
\xc{This approach incorporates semantic prototypes \citep{huang2025timedp} to extract implicit domain features from TS, complementing text-based conditioning that provides explicit domain information for diffusion model.} As a result, the proposed model achieves SOTA performance across multiple datasets on fidelity of the generation results and demonstrates controllability in both in-domain and out-of-domain settings.

\xc{
To summarize, this paper presents the following novel contributions:
\textbf{First}, we introduce a multi-agent framework for creating a text-controlled TSG dataset. Our numerical experiments show that textual descriptions provide valuable information for time-series models.
\textbf{Second}, using this dataset, we analyze the impact of different types of time-series descriptions, advancing the understanding of how LLMs can assist in time-series prediction and generation.
\textbf{Third}, we propose \frameworkname{}, a novel framework for text controlled TSG through diffusion models. Our approach excels in generating highly controllable time-series, outperforming baseline models across 11 out of 12 datasets, and highlights its potential for tackling complex, real-world tasks, with promising applications in healthcare, finance, and beyond.
}

\vspace{-0.1in}
\section{Related work}
\paragraph{Using Text for Time Series Modeling:}   Text-based approaches have shown promise in enhancing various time series tasks, including forecasting~\citep{DBLP:journals/corr/abs-2310-01728,DBLP:conf/nips/GruverFQW23,DBLP:journals/corr/abs-2306-03763,DBLP:journals/tkde/XueS24,zhang2024timeraf}, classification~\citep{xie2023wall,DBLP:journals/corr/abs-2304-07619}, and event prediction~\citep{DBLP:journals/corr/abs-2305-14847,shi2024language}. While these studies primarily focus on leveraging text to guide or interpret existing time series, text-to-time series generation remains underexplored. GenG~\citep{DBLP:conf/infocom/ZhouJHX0Y24} initiates this direction with a two-stage pipeline, but is limited to specific domains and lacks instance-level control. Time-MMD~\citep{DBLP:conf/nips/LiuXZKKSSCW0P24} presents a large-scale paired dataset for multi-domain text–time series forecasting. However, its step-level annotations focus on local transitions, limiting its applicability to modeling or evaluating alignment over extended temporal sequences.
 
\paragraph{Conditional Time Series Generation:}  Diffusion-based models have emerged as a powerful paradigm for conditional time series generation. A number of recent works employ denoising score matching and continuous-time diffusion to enable flexible, probabilistic generation~\citep{DBLP:conf/nips/TashiroSSE21, DBLP:conf/icml/ShenK23, DBLP:conf/icml/NarasimhanAASC24, DBLP:journals/corr/abs-2408-12991, kollovieh2024predict, DBLP:conf/iclr/ShenCK24, fan2024mg,deng2025tardiff, hou2024invdiff} These methods explore a range of network backbones, including score-based transformers~\citep{DBLP:conf/iclr/YuanQ24}, structured state space models~\citep{DBLP:journals/tmlr/AlcarazS23}, and constrained denoising objectives~\citep{DBLP:conf/nips/ColettaGBV23}. Some also incorporate seasonality and trend decomposition as explicit inductive biases~\citep{DBLP:conf/iclr/YuanQ24}. Despite their expressiveness, most models operate within single-domain scenarios and rely on fixed conditioning formats. One of the works with cross-domain focus is TimeDP \citep{huang2025timedp}, which introduces the use of prototypes to build `soft prompts'' for guiding cross-domain generation. However, its coarse-grained control makes it challenging to achieve personalized text-controlled TSG.

\section{Problem Formulation}\label{sec:pro}
\xc{
Time series often span multiple domains, each with unique temporal dynamics. Let $D = \{ D_1, D_2, \dots, D_k \}$ denote a set of domains, where each domain $D_i$ is associated with a collection of time series $X_i = \{\mathbf{x}_t \mid t = 1, 2, \dots, T\}$, and $\mathbf{x}_t \in \mathbb{R}^d$ represents a $d$-dimensional vector at timestamp $t$. 
Our aim is to generate realistic time series while capturing domain-specific patterns and adhering to constraints specified by textual descriptions. Formally, the goal is to learn a generative function:
$f_\theta: (D, l, z) \to \hat{\mathbf{x}}$,
where $l$ represents a textual prompt describing characteristics such as trends or periodicity, and $z$ is a latent variable sampled from a prior distribution. The output $\hat{\mathbf{x}}$ is a time series that aligns with both domain-specific patterns and textual conditions.
}

Adhering to the channel-independent setting~\citep{patchtst} that is widely accepted by recent researches, we formulate the problem studied in this paper in a uni-variate time series generation manner to handle the heterogeneity of time series in terms of dimension~\citep{moirai}.

\section{Methodology}

We propose a unified framework for text-controlled time series generation, which consists of two main stages: \textit{text-to-time series data preparation} and \textit{text-to-time series data generation}. 
The overall architecture is illustrated in Figure ~\ref{fig:overview}, where both stages are tightly coupled to facilitate high-quality, controlled time series synthesis.

\subsection{Text-to-Time Series Data Preparation}

To address the scarcity of high-quality text-to-time series paired datasets, we investigate how to generate effective textual descriptions of time series data to create high-quality TS-text-paired datasets.
Initially, we explored some straightforward methods, such as rule-based approaches that relied on simple trend-related terms  (e.g., ``increasing'', ``decreasing'') and degree modifiers (e.g., ``significant'', ``slight'').
Additionally, we tried to leverage GPT-4o and incorporated Seasonal-Trend Decomposition using Loess (STL) \citep{cleveland1990stl} to preprocess time-series data by decomposing it into trend, seasonal, and residual components, making the TS data easier for the model to interpret.
However, the resulting texts were overly simplistic and failed to capture the complex patterns and domain-specific nuances inherent in time-series data~(results in Appendix \ref{LLM_directly}). 
Therefore, we further explore leveraging more diverse and enriched sources to generate text capable of effectively assisting time-series tasks. 

\begin{figure*}[t]
    \centering
\includegraphics[scale=0.5]{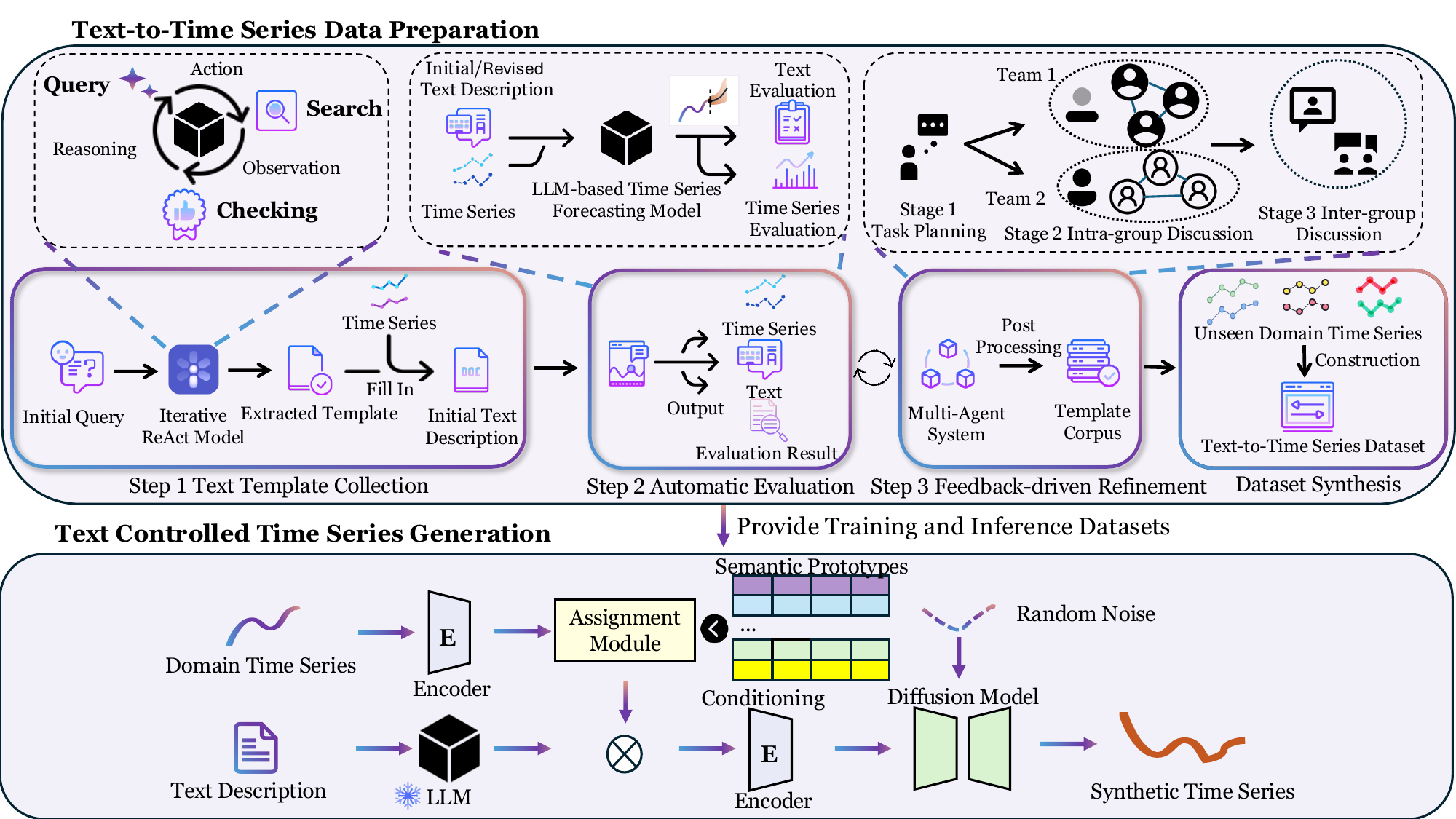}

\caption{Overview of the proposed \frameworkname{} framework, which consists of two main stages: \textit{text-to-time series data preparation} and \textit{text-to-time series data generation}. The former consists of three steps: \emph{(1)} collecting initial text templates from online sources, \emph{(2)} performing automatic evaluation to assess the quality of generated descriptions, and \emph{(3)} employing a feedback-driven refinement process via a multi-agent system to iteratively improve the textual outputs. The latter, i.e., the diffusion-based generation model, we adopt a hybrid strategy where \emph{(i)} a textual description is used as input. Meanwhile, \emph{(ii)} we utilize the target domain's time series and extract its corresponding domain prototypes and their weights using Prototype Assignment Module. These prototypes and weights are also used as inputs and are fused with the textual description.
This fusion serves as the conditioning input for the diffusion model, which is applied to perform text-controlled TSG.}

    \label{fig:overview}
\end{figure*}

To this end, We propose a multi-agent system to automatically generate and iteratively refine TS textual descriptions. 
As shown in Figure~\ref{fig:overview}, the proposed framework comprises three key components: 
\textit{Step 1: Text Template Generation}, which focuses on collecting and extracting text templates to construct initial textual descriptions; 
\textit{Step 2: Automated Evaluation}, designed to assess the effectiveness of the descriptions in supporting downstream tasks; and 
\textit{Step 3:Feedback-Driven Refinement}, which improves the textual descriptions based on evaluation metrics. Steps 2 and 3 will alternate and iterate until the agent system determines the output is sufficiently refined or a predefined iteration limit is reached.
Throughout the iterations, the agents refine a set of general-purpose text templates, which are designed to be dataset-agnostic and easily adaptable to new domains and datasets.
Subsequently, these refined templates are utilized to generate textual descriptions for TSG tasks.

\textbf{Step 1 \xc{Text Template Collection}:} 
As noted in previous work~\citep{DBLP:journals/corr/abs-2404-11757}, generating fine-grained text descriptions remains a challenging task due to the limited availability of extensive data resources, \textcolor{Violet}{while also avoiding the potential information leakage that could occur when interacting with external sources while generating text descriptions for a single data. 
To address this limitation, we adopt a template-based approach that standardizes the narrative of key TS information.} Starting with a variety of initial queries, we first collect articles, news, and reports that \xc{describe TS data}. Inspired by ReAct~\citep{DBLP:conf/iclr/YaoZYDSN023}, we propose a single-agent framework, which prompts LLMs to generate dynamic reasoning traces for collecting candidates and actions to interact with external environments (e.g., Google, Wikipedia) in an interleaved manner~\citep{DBLP:conf/nips/MadaanTGHGW0DPY23} (Framework pipeline can be find in Appendix~\ref{text_single-agent}). 
The agent decomposes the query into sub-questions, using external tools to answer each sub-question iteratively until all are addressed. 
\textcolor{Violet}{Afterwards, another LLM extracts general TS templates from the collected documents, thus curating a set of 50 general-purpose templates.
To ensure broad applicability across domains, dataset-specific details are carefully excluded through a combination of prompting techniques and human verification.
During the dataset construction phase, an LLM is employed to fill these templates with domain-specific and time-series-related details, resulting in initial textual descriptions tailored to the target domain. The detailed example can be viewed in Appendix~\ref{template_bank}.}

\xc{
\textbf{Step 2 Automatic Evaluation:} 
Our goal is to equip the system with the ability to evaluate generated text and provide iterative feedback, aligning with the broader objective of leveraging text to guide TSG.
This is achieved by incorporating a ts task conditioned on accompanying text, based on the premise that higher-quality text improves performance.
While a straightforward approach might involve pre-training or fine-tuning existing models specifically for TSG, this would demand substantial computational resources. 
Such an approach would be highly resource-intensive, as it would require training or fine-tuning at every iteration.  }

\xc{Because recent advancements in prompting strategies 
have demonstrated the potential
of LLMs for zero-shot Time Series Forecasting (TSF)~\citep{DBLP:conf/nips/GruverFQW23}, we use LSTPrompt~\citep{DBLP:conf/acl/LiuZWKP24} and LLMTime~\citep{DBLP:conf/nips/GruverFQW23}
as our evaluation backbone for a more compute-efficient alternative to model training. 
Specifically, we prompt off-the-shelf LLMs with chain-of-thought (CoT) reasoning~\citep{DBLP:conf/nips/Wei0SBIXCLZ22}, which enables the integration of text as an additional input modality~\citep{DBLP:conf/acl/LiuZWKP24}.  
For the detailed automatic evaluation pipeline, initial evaluation criteria and definitions, refer to Appendix~\ref{automatic_evaluation} and \ref{dataset_evaluation}.
Additionally, we conducted evaluations on the impact of different text types on TSG---results shown in Appendix~\ref{texttype_diffusion} demonstrate that generated texts that improve  LSTPrompt performance in TSF, also enhances TSG performance.}

\textbf{Step 3 Feedback-driven Refinement:}  
\xc{Based on the initial text templates and the automatic evaluation system, 
we further} propose a multi-agent collaboration system that simulates the iterative refinement process of a team of human prompt engineers, leveraging the demonstrated capability of LLMs to improve their own outputs \citep{zhang2024can}. 
As illustrated in Figure \ref{fig:overview}, the system operates through three stages: \textbf{Stage 1 Task Planning} (assigning tasks and monitoring progress), \textcolor{Violet}{a manager agent orchestrates the workflow, laying the groundwork for all subsequent stages. This agent assigns specific tasks to independent teams, monitors progress, and ensures alignment across iterations.} \textbf{Stage 2 Intra-group Discussion} (independent teams iteratively refining outputs), \textcolor{Violet}{two independent teams work iteratively to refine the text. Each team includes agents with specific roles: a planner who coordinates tasks, a scientist who analyzes and identifies improvements, an engineer who implements changes, and an observer who critiques the outputs. The teams refine their work through cycles until a satisfactory output is produced.}  \textbf{Stage 3 Inter-group Discussion}  (collaborative consensus building), \textcolor{Violet}{the two teams’ leaders, moderated by the manager, engage in a structured dialogue to compare and integrate their results. This collaborative process continues until a consensus is reached, producing a unified output. Once all initial templates are refined, the system extracts them into a new set of templates in \textbf{Stage 4 Post Processing}. 
These final templates are filtered to remove duplicates and ensure they do not contain any dataset-specific or time-series-specific information, resulting in a generalized template library. }
\textcolor{Violet}{This library is fixed and remains unaltered during subsequent dataset construction, ensuring consistency and reusability. Further details about the system structure, processing pipeline, and output examples can be found in Appendices~\ref{appendix_system_pipeline}, \ref{text_dimensions} and \ref{template_list}.} 

\textbf{Dataset Synthesis:}
\xc{
Excluding domain- or dataset-specific information during template construction ensures their generalizability to unseen domains. 
To demonstrate this, templates generated with the methodology described in the previous subsection used only two datasets and are applied to 12 additional, entirely disjoint datasets in our TSG experiments (seen Section~\ref{7.2}).
During the dataset construction process, information specific to individual time series is extracted, while domain-specific details, which are pre-defined in the dataset, remain consistent across all instances. 
Specifically, a standalone LLM, separate from the multi-agent framework, is responsible for extracting statistical information from the time series and populating the templates accordingly. 
This process is entirely offline, with no reliance on external networks or the LLM’s pre-trained knowledge. 
Once populated, the dataset remain static and unchanged throughout the TSG phase.
}

\subsection{Text Controlled Time Series Generation}

\xc{
Although text provides explicit details on trends, statistical properties, and domain-specific information, its discrete nature contrasts with the continuous structure of time-series data, posing challenges for using text as a control mechanism. 
To address this, we present our framework 
for text-controlled TSG. As shown in Figure~\ref{fig:overview}, the framework employs a hybrid prompt that integrates \textit{semantic prototypes}
which supplement coarse-grained domain descriptions provided by text, along with textual descriptions.
This design captures shared patterns across domains and improves the model's generalization ability in unseen domains. }

\xc{We selected diffusion models \citep{ho2020denoising,lin2024diffusion} as the backbone of our framework due to their proven ability to generate diverse, high-quality data while effectively capturing complex data distributions. 
Recent successes of diffusion models in time-series forecasting \citep{rasul2021autoregressive} and generation \citep{qian2024timeldm} further underscore their suitability for this task. 
The specific design of our architecture is detailed in Appendix \ref{algorithm}.
}

\subsubsection{Domain-specific Prototype Matching}

\xc{
Relying solely on text descriptions often leads to insufficient representation of domain-specific patterns due to their coarse-grained and abstract nature. 
To address this, we draw inspiration from TimeDP~\citep{huang2025timedp}, which introduces an automated approach to extracting TS domain features.
We propose a hybrid representation strategy that complements domain information using semantic prototypes. 
Semantic prototypes serve as ``bases'', representing elementary features of time series such as trends and seasonality. 
These bases act as a shared ``dictionary'' across different domains, with each prototype vector serving as a ``word'' that encodes specific semantic features of TS data.
}

\xc{Specifically, we introduce a set of vectors as time series prototypes $\bcP\in \mathbb{R}^{N_p\times d}$ for representing cross-domain time series common knowledge, where each prototype vector $\bsp \in \mathbb{R}^{1\times d}$ serves as the representation of a time series basis and $N_p$ is the number of prototypes.  
Each time series sample corresponds to a distinct allocation of these bases. 
The mapping from time-series samples to their respective allocations highlights the importance of prototypes for individual instances and distinguishes among domains. 
We propose \textit{Prototype Assignment Module} to extract domain-specific prototype weights $\bsm$, with domain characteristics expressed through the selection and weighting of these bases.
Then, the extracted prototypes $\bcP$ and their corresponding weights $\bsm$, denoted as $(\bcP, \bsm)$, are leveraged as supplementary information to enhance the text domain representation, enriching its semantic understanding and cross-domain adaptability. 
During inference, samples from the target domain are used to extract prototypes and compute weights. }


\subsubsection{Model Training}
We integrate the semantic prototypes $\bsp$ with weights $\bsm$ and the embeddings of textual descriptions $l$ to construct a hybrid prompt. This prompt is then fed into the cross-attention layers of diffusion-based generation model, enhancing the expressiveness of textual inputs.  
This enables precise and effective controlled generation.  
With the conditional denoising mechanism, the denoising objective using $\epsilon$-parameterization~\citep{ho2020denoising} can be written and simplified as:

\begin{align}
    L &= \mathbb{E}[\|\bep-\hat{\bep}\|^2] \nonumber\\
    &= \mathbb{E}_{\bsx_0 \in {D^T},{\bep}\sim \mathcal{N}(\mathbf 0,\mathbf I),n}[\|\bep-\bep_{\theta,\bsP}({x}_n,n,\bsm, \bsl)\|^2]
    \label{eq:condloss}
\end{align}

where $n$ denotes the denoising step.  
Please refer to Section~\ref{sec:pro} for symbol definitions. For implementation details, see Appendix~\ref{Semantic_Prototype_Assignment_Alignment}.

\section{Experiment Setup and Result Analysis}

\textcolor{Violet}{
Our experiments are designed to investigate the key claims of this paper: The feasibility of using text for TSG with \frameworkname{}, its generalisability to unseen domains (Section~\ref{7.2}), and its capability to provide instance-level control in both in-domain and out-of-domain scenarios (Section~\ref{7.4}). We also conduct a comprehensive analysis of textual description types and agent strategies (Section~\ref{7.1} and \ref{7.7}), as well as investigate the impact of \frameworkname{}’s parameter choices and configuration settings (Section~\ref{7.6}).}

\subsection{Experimental Setup}

\subsubsection{Baselines}
We select SOTA methods for both TS generation and forecasting tasks as baselines. For generation, we explore the performance of \frameworkname{} by comparing with conditional (\texttt{TimeVQVAE},~\citealt{DBLP:conf/aistats/LeeMA23}) and unconditional approaches~(\texttt{TimegGAN},~\citealt{DBLP:conf/nips/YoonJS19}; \texttt{GT-GAN},~\citealt{DBLP:conf/nips/JeonKSCP22}; \texttt{TimeVAE},~\citealt{DBLP:journals/corr/abs-2111-08095}). For forecasting, our goal is to establish the realism of synthetic data. Here, we compare the performance of \texttt{Time-LLM} \citep{DBLP:journals/corr/abs-2310-01728}, \texttt{LLM4TS} \citep{DBLP:journals/corr/abs-2308-08469} and \texttt{TEMPO} \citep{DBLP:conf/iclr/CaoJAPZY024}, \texttt{GPT4TS} \citep{DBLP:conf/nips/ZhouNW0023}. Detailed descriptions can be found in Appendix \ref{TS_forecasting_model}. More details about Experiment Setup and Implementation can be found at Appendix~\ref{experiment_setup} and Appendix \ref{implementation_detail}.

\subsubsection{Datasets}
\textcolor{Violet}{We employ AirPassenger and Sunspots as benchmark datasets for multi-agent system assessing the text types impact.} We evaluate the effectiveness of \frameworkname{} on 12 in-domain datasets including Electricity, Solar, Wind, Traffic, Taxi, Pedestrian, Air, Temperature, Rain, NN5, Fred-MD, Exchange. These datasets have been widely used as benchmark datasets for TSG tasks and obtain from GluonTS \citep{gluonts_jmlr} and Monash Time Series Forecasting Repository \citep{godahewa2021monash}.  \textcolor{Violet}{For the evaluation of unknown domains, we selected the stock and web dataset.} Finally, We use ILI and M4 \citep{makridakis2018m4} \textcolor{Violet}{as additional datasets for Time Series Forecasting task.(Details shown in Appendix~\ref{statistics_dataset}.)} 

\subsubsection{Evaluation Metrics}

\textbf{Assessing Fidelity of TSG:} To evaluate the impact of text types, we use the Mean Absolute Error (MAE). For TC-TSG, we measure the Marginal Distribution Discrepancy (MDD) and Kullback-Leibler (K-L) divergence to quantify the realism of the synthesized data, with further details provided in Appendix \ref{baseline_evaluation}. 

\textbf{Assessing Controllability of TSG:} To evaluate controllability in text-to-time series generation (TSG), we adopt the \textbf{Joint Fréchet Time Series Distance (J-FTSD)} \citep{DBLP:conf/icml/NarasimhanAASC24}, which captures both local shape and global distributional similarity between the generated and reference sequences. This metric offers a more faithful measure of alignment than traditional point-wise metrics such as MSE or MAE, especially in cases where semantic similarity does not require exact numerical overlap. However, quantitative metrics alone may overlook perceptual or contextual alignment. To complement this, we conduct a \textbf{Human Evaluation}, where outputs are ranked based on how well they reflect the intended text descriptions. We consider two settings: in HE-Rank, annotators evaluate only the outputs generated by different model settings along with the ground truth time series; in HE-Mixed, these candidates are presented alongside three randomly selected distractor time series from the test set. This design enables both focused and distractor-aware evaluation, capturing subtle differences (e.g., constant shifts) while assessing trend or pattern alignment. Results are reported as average ranks across multiple evaluations. See Appendix~\ref{human_evaluation} for details.

\textbf{Time Series Forecasting:}
Following previous work~\citep{DBLP:conf/iclr/WuHLZ0L23}, we measure the MSE and MAE for long-term forecasting. For short-term forecasting on the M4 benchmark, we adopt Symmetric Mean Absolute Percentage Error (SMAPE), Mean Absolute Scaled Error (MASE), and Overall Weighted Average (OWA) as evaluation metrics \citep{DBLP:conf/iclr/OreshkinCCB20}.

\begin{table*}[t]
    \centering
    \renewcommand{\textcolor}[2]{\keepcolor{#1}{#2}} 
    \begin{adjustbox}{width=\textwidth}
    \begin{tabular}{llccccccc}
    \toprule
      & \textbf{Dataset}  & \textbf{\frameworkname{}} & \textbf{\frameworkname{} w/o Text}  & \textbf{\frameworkname{} w/o Prototype} & \textbf{TimeVQVAE} & \textbf{TimeGAN} & \textbf{GT-GAN} & \textbf{TimeVAE}\\ 
    \midrule
      \multirow{12}{*}{\rotatebox[origin=c]{90}{\footnotesize{\textbf{Marginal Distribution Distance}}}} & Electricity  & \textcolor{blue}{$0.220 \pm 0.070$} & \textcolor{red}{$0.202 \pm 0.066$} & $0.277 \pm 0.068$ &$1.763 \pm 0.088$  & $2.443 \pm 0.765$ & $2.026 \pm 0.280$ & $3.306 \pm 0.044$\\
      
      & Solar & \textcolor{red}{$375.531 \pm 0.001$} & $375.908 \pm 10.230$  & $376.111 \pm 5.506$  & $466.174 \pm 0.145$ & $460.810 \pm 14.078$ & $476.196 \pm 17.041$ & \textcolor{blue}{$365.906 \pm 6.365$} \\ 
      
      & Wind & \textcolor{red}{$0.316 \pm 0.031$} & \textcolor{blue}{$0.319 \pm 0.046$}  & $0.362 \pm 0.050$  & $0.777 \pm 0.028$  & $1.115 \pm 0.159$ & $0.706 \pm 0.106$ & $0.943 \pm 0.008$ \\
      
      & Traffic & \textcolor{red}{$0.254 \pm 0.034$} & \textcolor{blue}{$0.261 \pm 0.038$}  & $0.316 \pm 0.006$  & $1.170 \pm 0.028$ & $1.733 \pm 0.137$ & $1.311 \pm 0.032$ & $0.984 \pm 0.012$ \\
      
      & Taxi & \textcolor{red}{$0.386 \pm 0.057$} & \textcolor{blue}{$0.391 \pm 0.057$}  & $0.491 \pm 0.133$  & $0.534 \pm 0.032$ & $1.278 \pm 0.168$ & $1.118 \pm 0.157$ & $0.697 \pm 0.007$ \\
      
      & Pedestrian & \textcolor{red}{$0.621 \pm 0.124$} & \textcolor{blue}{$0.624 \pm 0.116$}  & $0.800 \pm 0.111$ & $1.225 \pm 0.060$ & $1.574 \pm 0.290$ & $1.559 \pm 0.117$ & $0.777 \pm 0.012$ \\
      
      & Air & $0.447 \pm 0.112$ & \textcolor{blue}{$0.410 \pm 0.129$}  & $0.508 \pm 0.091$  & \textcolor{red}{$0.338 \pm 0.012$} & $2.089 \pm 0.618$ & $2.828 \pm 0.172$ & $1.369 \pm 0.040$ \\
      
      & Temperature & \textcolor{red}{$0.342 \pm 0.010$} & \textcolor{blue}{$0.345 \pm 0.019$}  & $0.408 \pm 0.053$  & $0.943 \pm 0.035$ & $1.164 \pm 0.110$ & $1.165 \pm 0.072$ & $2.044 \pm 0.024$ \\
      
      & Rain & \textcolor{red}{$5.340 \pm 0.421$} & \textcolor{blue}{$5.597 \pm 0.409$}  & $6.678 \pm 2.045$ & $9.243 \pm 0.122$ & $10.937 \pm 4.039$ & $6.473 \pm 1.207$ & $9.134 \pm 0.477$ \\
      
      & NN5 & \textcolor{red}{$0.591 \pm 0.029$} & \textcolor{blue}{$0.628 \pm 0.021$} & $0.748 \pm 0.134$ & $1.424 \pm 0.043$ & $2.758 \pm 0.142$ & $2.121 \pm 0.094$ & $2.871 \pm 0.045$ \\
      
      & Fred-MD & \textcolor{red}{$0.258 \pm 0.045$} & \textcolor{blue}{$0.271 \pm 0.056$}  & $0.359 \pm 0.097$  & $2.932 \pm 0.133$ & $4.028 \pm 0.130$ & $4.026 \pm 0.087$ & $2.902 \pm 0.215$ \\
      
      & Exchange & \textcolor{red}{$0.374 \pm 0.053$} & \textcolor{blue}{$0.376 \pm 0.058$}  & $0.382 \pm 0.041$  & $0.993 \pm 0.058$ & $1.553 \pm 0.122$ & $1.355 \pm 0.072$ & $1.331 \pm 0.042$ \\
      \midrule
      \multirow{12}{*}{\rotatebox[origin=c]{90}{\footnotesize{\textbf{K-L Divergence}}}} & Electricity & \textcolor{red}{$0.011 \pm 0.010$} & $0.014 \pm 0.013$  & \textcolor{blue}{$0.013 \pm 0.005$}  & $0.185 \pm 0.018$ & $0.395 \pm 0.121$ & $0.415 \pm 0.040$ & $0.580 \pm 0.005$ \\
      
      & Solar & \textcolor{red}{$0.007 \pm 0.002$} & \textcolor{blue}{$0.007 \pm 0.003$}  & $0.015 \pm 0.012$  & $0.726 \pm 0.043$ & $0.889 \pm 0.288$ & $0.102 \pm 0.045$ & $0.201 \pm 0.008$ \\
      
      & Wind & \textcolor{blue}{$0.067 \pm 0.030$} & \textcolor{red}{$0.061 \pm 0.042$}  & $0.069 \pm 0.030$ & $0.493 \pm 0.081$ & $4.528 \pm 1.743$ & $0.511 \pm 0.129$ & $0.553 \pm 0.014$ \\
      
      & Traffic & \textcolor{red}{$0.013 \pm 0.004$} & \textcolor{blue}{$0.013 \pm 0.005$}  & $0.017 \pm 0.004$ & $0.145 \pm 0.015$ & $2.134 \pm 0.952$ & $1.108 \pm 0.171$ & $0.212 \pm 0.006$ \\
      
      & Taxi & \textcolor{red}{$0.013 \pm 0.009$} & \textcolor{blue}{$0.013 \pm 0.010$}  & $0.039 \pm 0.031$ & $0.100 \pm 0.014$ & $1.160 \pm 0.651$ & $0.663 \pm 0.127$ & $0.120 \pm 0.005$ \\
      
      & Pedestrian & \textcolor{red}{$0.011 \pm 0.009$} & \textcolor{blue}{$0.011 \pm 0.009$} & $0.020 \pm 0.005$  &  $0.275 \pm 0.021$ & $0.881 \pm 0.436$ & $0.347 \pm 0.085$ &  $0.052 \pm 0.010$ \\
      
      & Air & $0.022 \pm 0.017$ & \textcolor{blue}{$0.018 \pm 0.014$} & $0.019 \pm 0.011$ & \textcolor{red}{$0.017 \pm 0.004$} & $0.588 \pm 0.369$ & $0.506 \pm 0.091$ & $0.176 \pm 0.016$ \\
      
      & Temperature &  \textcolor{red}{$0.023 \pm 0.013$} &  \textcolor{blue}{$0.024 \pm 0.010$}  & $0.039 \pm 0.033$ & $0.980 \pm 0.190$ & $8.775 \pm 2.511$ & $2.177 \pm 0.323$ & $1.910 \pm 0.076$ \\
      
      & Rain &\textcolor{red}{$0.006 \pm 0.001$} & \textcolor{blue}{$0.007 \pm 0.001$}  & $0.009 \pm 0.003$ & $0.008 \pm 0.002$ & $0.383 \pm 0.089$ & $0.462 \pm 0.056$ & $0.175 \pm 0.011$ \\
      
      & NN5 & \textcolor{red}{$0.010 \pm 0.008$} & \textcolor{blue}{$0.013 \pm 0.008$}  & $0.016 \pm 0.007$ & $0.603 \pm 0.107$ & $4.054 \pm 1.592$ & $1.372 \pm 0.180$ & $1.284 \pm 0.058$ \\
      
      & Fred-MD & \textcolor{red}{$0.024 \pm 0.019$} & \textcolor{blue}{$0.028 \pm 0.029$} & $0.033 \pm 0.029$  & $0.712 \pm 0.054$ & $5.371 \pm 1.455$ & $3.509 \pm 0.299$ & $0.376 \pm 0.025$ \\
      
      & Exchange & \textcolor{red}{$0.083 \pm 0.056$} & \textcolor{blue}{$0.088 \pm 0.065$}  & $0.088 \pm 0.077$ & $1.984 \pm 0.836$ & $4.376 \pm 0.664$ & $1.583 \pm 0.932$ & $2.011 \pm 0.433$ \\
      
    \bottomrule
    \end{tabular}
    \end{adjustbox}
    \renewcommand{\textcolor}[2]{#2}
    \caption{Generation results on various univariate datasets, evaluated using Marginal Distribution Distance (MDD) and K-L divergence (K-L). Lower values indicate better performance. Best results are in \textcolor{red}{red}, and second best in \textcolor{blue}{blue}.}
    \label{tab:mdd_kl_main}
\end{table*}

\subsection{Overall Generation Performance on Fidelity}
\label{7.2}
\subsubsection{Performance on in-domain setting}
\label{6.4.1}
\textcolor{Violet}{\textbf{Objective:} We evaluated the overall fidelity of \frameworkname{} using 12 in-domain datasets. During inference, the text from the test set served as prompts, where the prototypes are generated by the training set.}

\textcolor{Violet}{As shown in Table \ref{tab:mdd_kl_main}, \frameworkname{} consistently outperforms existing baselines across a variety of datasets, demonstrating its robustness and versatility. In terms of MDD, \frameworkname{} (w/o Text) achieves the best performance on most datasets, ranking second only on the Air dataset. For instance, on the Wind dataset, \frameworkname{}  achieves an MDD of $0.316 \pm 0.031$, significantly outperforming models like TimeVAE ($0.943 \pm 0.008$) and TimeGAN ($1.115 \pm 0.159$).  The KL divergence results further underscore \frameworkname{}'s capabilities. \frameworkname{} achieves the lowest KL divergence on almost all datasets, ranking second only on the Wind and Air datasets. On the Electricity dataset, \frameworkname{} achieves a KL divergence of $0.011 \pm 0.010$, substantially better than \frameworkname{} (w/o Text) at $0.014 \pm 0.013$, and significantly outperforming baselines such as TimeVQVAE ($0.185 \pm 0.018$) and TimeGAN ($0.395 \pm 0.121$). Notably, even without text conditioning, \frameworkname{} often secures the second-best performance, highlighting the strength of its core architecture. For example, on the Pedestrian dataset, \frameworkname{} (w/o Text) achieves the second-best KL divergence of $0.011 \pm 0.009$, only slightly behind the top-performing model.} \textcolor{Violet}{ We also show that \frameworkname{} can generate realistic synthetic data for forecasting downstream tasks in Appendix~\ref{downstream}.}

\subsubsection{Performance on Unseen domain settings}
\label{7.3}

\textcolor{Violet}{\textbf{Objective:} We evaluated the performance of \frameworkname{} in unseen domains, training on 12 in-domain datasets and testing on two out-of-domain datasets. In this setting, the few-shot scenario established the prototypes, while the text functioned as the control condition for generation .}

\begin{table}[htbp]
    \centering    
        \begin{adjustbox}{width=0.9\columnwidth}
            \begin{tabular}{llcccc}
                \toprule
                & \multirow{2}{*}{\textbf{Methods}} & \multicolumn{2}{c}{\textbf{MDD}} & \multicolumn{2}{c}{\textbf{K-L}}\\
                \cmidrule(lr){3-4} \cmidrule(lr){5-6}
                & {} & 5-shots & 10-shots & 5-shots & 10-shots \\
                \midrule
                \multirow{5}{*}{\rotatebox[origin=c]{90}{\footnotesize{\textbf{Stock}}}} & TimeVQVAE & $3.502$ & $3.514$ & \underline{$1.487$} & $3.785$\\
                & TimeGAN & $3.834$ & $3.765$ & $14.347$ & $13.823$\\
                & GT-GAN & $3.653$ & $3.474$ & $10.971$ & $8.855$\\
                & TimeVAE & $3.738$ & $3.338$ & $6.048$ & $4.479$\\
                & \frameworkname{} & \underline{$3.477$} & \underline{$3.112$} & $3.249$ & \underline{$2.827$} \\
                \midrule
                \multirow{5}{*}{\rotatebox[origin=c]{90}{\footnotesize{\textbf{Web}}}} & TimeVQVAE & $9.630$ & $10.012$ & $1.665$ & $1.213$\\
                & TimeGAN & $8.304$ & $8.122$ & $4.343$ & $3.236$\\
                & GT-GAN & \underline{$8.018$} & $8.936$ & $10.037$ & $10.915$\\
                & TimeVAE & $10.106$ & $10.211$ & $3.332$ & $1.845$\\
                & \frameworkname{} & $8.085$ & \underline{$7.995$} & \underline{$0.905$} & \underline{$0.876$}\\
                \bottomrule
            \end{tabular}
        \end{adjustbox}
        \caption{Few-shot Performance of Unseen domain. We compare the proposed methods and baseline on 5,10-shots. Best results are highlighted in bold face.}
        \label{tab:few-shot}
\end{table}

Our model demonstrates superior robustness in both MDD and K-L metrics, outperforming the baseline models in most cases (Shown in Table~\ref{tab:few-shot}). Specifically, it achieves the best results for general MDD and K-L scores at both 5-shot and 10-shot settings. Furthermore, the performance improvement with additional examples suggests that the model effectively leverages learned semantic prototypes to recall more accurate domain and pattern information, enhancing its generalization capability.

\subsection{Controlling Generation Performance}
\label{7.4}
\subsubsection{Performance on In-domain settings}
\textcolor{Violet}{\textbf{Objective:} We evaluated the controllability of \frameworkname{} on 12 in-domain datasets using J-FTSD \citep{DBLP:conf/icml/NarasimhanAASC24} and human evaluation, while keeping all other settings consistent with Subsection \ref{6.4.1}.}

\begin{table*}[t]
    \centering    
    \begin{adjustbox}{width=\textwidth}
    \begin{tabular}{lccc ccc ccc ccc}
        \toprule
        & \multicolumn{3}{c}{\textbf{Electricity}} & \multicolumn{3}{c}{\textbf{Solar}} & \multicolumn{3}{c}{\textbf{Wind}} & \multicolumn{3}{c}{\textbf{Traffic}} \\
         &J-FTSD & HE-Rank & HE-Mixed & J-FTSD & HE-Rank & HE-Mixed & J-FTSD & HE-Rank & HE-Mixed & J-FTSD & HE-Rank & HE-Mixed \\
        \cmidrule(lr){2-4} \cmidrule(lr){5-7} \cmidrule(lr){8-10} \cmidrule(lr){11-13}
        \frameworkname{}  & $0.538$ & $2.3$ & $2.4$ & $0.295$ & $2.6$ & $2.8$ & $5.011$ & $1.8$ & $1.8$ & $0.570$ & $1.4$  & $1.4$ \\
        \textit{w/o Prototype}  & $1.164$ & $3.3$ & $3.4$ & $0.322$ & $3.2$ & $3.3$ & $6.843$ & $3.1$ & $3.2$ & $0.597$ & $2.1$ & $2.1$\\
        \textit{w/o Text}  & $1.821$ & $3.4$ & $3.5$ & $0.330$ & $3.2$ & $3.4$ & $6.935$ & $4.0$ & $4.2$ & $0.611$ & $4.2$ & $4.4$\\
    
        \midrule
        & \multicolumn{3}{c}{\textbf{Taxi}} & \multicolumn{3}{c}{\textbf{Pedestrian}}
        & \multicolumn{3}{c}{\textbf{Air}} & \multicolumn{3}{c}{\textbf{Temperature}} \\
        &J-FTSD & HE-Rank & HE-Mixed & J-FTSD & HE-Rank & HE-Mixed & J-FTSD & HE-Rank & HE-Mixed & J-FTSD & HE-Rank & HE-Mixed \\
        \cmidrule(lr){2-4} \cmidrule(lr){5-7} \cmidrule(lr){8-10} \cmidrule(lr){11-13}
        \frameworkname{}  & $0.974$ & $1.4$ & $1.4$  & $0.488$ & $2.2$ & $2.6$ & $0.654$ & $2.7$ & $2.8$ & $3.977$ & $2.4$ & $2.4$\\ 
        \textit{w/o Prototype}  & $1.037$ & $3.4$ & $3.6$ & $0.662$ & $2.6$ & $3.1$ & $0.817$ & $3.0$ & $3.1$ & $4.613$ & $2.6$ & $3.1$ \\
        \textit{w/o Text} & $1.312$ & $4.2$ & $4.4$ & $0.550$ & $3.3$ & $4.0$ & $0.677$ & $3.3$ & $3.4$ & $4.708$ & $3.6$ & $3.9$ \\
        
        \midrule
        & \multicolumn{3}{c}{\textbf{Rain}} & \multicolumn{3}{c}{\textbf{NN5}} & \multicolumn{3}{c}{\textbf{Fred-MD}} & \multicolumn{3}{c}{\textbf{Exchange}}\\
        &J-FTSD & HE-Rank & HE-Mixed & J-FTSD & HE-Rank & HE-Mixed & J-FTSD & HE-Rank & HE-Mixed & J-FTSD & HE-Rank & HE-Mixed \\
        \cmidrule(lr){2-4} \cmidrule(lr){5-7} \cmidrule(lr){8-10} \cmidrule(lr){11-13}
        \frameworkname{} & $0.147$ & $1.9$ & $1.9$ & $0.972$ & $2.5$ & $2.8$ & $0.260$ & $2.5$ & $2.8$ & $1.581$ & $1.9$ & $1.9$\\
        \textit{w/o Prototype} & $0.141$ & $3.2$ & $3.2$ & $1.115$ & $3.1$ & $3.5$ & $0.341$ & $2.7$ & $3.0$ & $1.846$ & $3.1$ & $3.1$\\
        \textit{w/o Text}  & $0.151$ & $3.7$ & $3.7$ & $1.192$ & $3.4$ & $3.8$ & $0.423$ & $3.8$ & $4.2$ & $1.738$ & $3.7$ & $3.7$ \\
        \bottomrule
    \end{tabular}
    \end{adjustbox}
    \caption{Text control performance of in-domain settings. Measured by J-FTSD and human evaluation.}
    \label{tab:text_control}
\end{table*}

As shown in Table~\ref{tab:text_control}, the proposed \frameworkname{} consistently outperforms the two ablated variants in most of datasets, according to both the J-FTSD score and human evaluation metrics. The removal of textual input leads to the most severe performance drop, particularly in human assessments—demonstrating that text is essential for producing outputs that align with intended semantics. Without textual guidance, the model tends to generate less coherent and less interpretable sequences. For example, on the Traffic and Pedestrian datasets, HE scores increase by over 3 points without text, indicating poor alignment. The ``w/o Prototype'' setting shows moderate degradation, suggesting that while prototype representations enhance fine-grained alignment and structure, they are less critical than textual descriptions. Notably, the gap is especially visible on datasets with high temporal variability such as Wind and Temperature. These results confirm that both components—text and prototypes—play complementary roles in generating realistic and controllable time series.

\subsubsection{Performance on Unseen domain settings}
\label{7.5}
\textcolor{Violet}{\textbf{Objective:} Same to the in-domain setting, we evaluated the controllability of \frameworkname{} on 2 unseen domain datasets.}

Table~\ref{tab:text_control_outofdomain} highlights the structural contributions of text and prototype components in unseen domains. The full \frameworkname{} model achieves the lowest J-FTSD and most favorable human evaluation scores across both Stock and Web datasets, indicating strong generalization beyond the training distribution. Removing text input causes the most notable performance drop, especially in J-FTSD and HE, confirming that textual guidance is critical for semantic alignment and coherent generation. In comparison, the “w/o Prototype” variant shows only moderate degradation, particularly in HE but not in J-FTSD, suggesting that prototypes help enforce structural consistency but are less essential than text for global control. These results support a complementary division of labor: text governs high-level semantics, while prototypes refine local dynamics.

\begin{table}[htbp]
    \centering    
    \begin{adjustbox}{width=\columnwidth}
    \begin{tabular}{lcccccc}
        \toprule
        & \multicolumn{3}{c}{\textbf{Stock}} & \multicolumn{3}{c}{\textbf{Web}} \\
         & J-FTSD & HE-Rank & HE-Mixed & J-FTSD & HE-Rank & HE-Mixed \\
        \cmidrule(lr){2-4} \cmidrule(lr){5-7} 
        \frameworkname{} & $7.483$ & $2.8$ & $2.8$ & $5.529$ & $2.7$ & $3.2$\\
        \textit{w/o Prototype} & $7.687$ & $2.8$ & $3.5$ & $5.752$ & $3.1$ & $3.5$\\
        \textit{w/o Text} & $8.178$ & $3.4$ & $4.0$ & $6.302$ & $3.2$ & $3.4$\\
        \bottomrule
    \end{tabular}
    \end{adjustbox}
    \caption{Comparison of text control performance of different settings on unseen domains. Measured by J-FTSD and human evaluation.}
    \label{tab:text_control_outofdomain}
\end{table}

\subsection{Performance Analysis on Controlling Text}
\label{7.1}

\textcolor{Violet}{\textbf{Objective:} We conducted a comprehensive analysis of the influence of text types. Specifically, we performed TSF experiments on two benchmark datasets to assess the impact of different text types.}

\textbf{Conciseness leads to better performance.} Table~\ref{tab:zeroshot_testing_strategy} shows that concise text inputs outperform overly detailed ones, which can mislead the model. This is particularly evident in the case of w/o instance context'', where the MAE improves by $1.6$ (compared to Initial text'') on the AirPassenger dataset, indicating that generating text that fully aligns with human preferences remains a challenging task. Notably, when it comes to longer sequence length, the context provides more useful information ($48.64$ vs $59.91$ on Sunspots).
\textbf{Clearly specifying the length of the prediction/generation can make the model's performance more stable.} This can be seen from the performance of ``w/o statistics''. After providing a clear sequence length and statistical values, the model's performance improves.
\textbf{Background information helps the model.} Similar to the findings of other works~\citep{DBLP:journals/corr/abs-2310-01728, DBLP:journals/corr/abs-2404-11757}, background information can significantly improve the model's performance. This is likely because retrieving the pre-trained knowledge from the LLMs can offer additional contextual information as support.
\textbf{Direct pattern descriptions are more effective than detailed trend descriptions.} As mentioned in Appendix~\ref{LLM_directly}, when attempting to decompose the TS into seasonal, trend, and residual components, the model's performance did not show significant improvement. After multiple iterations, the most effective method was to provide the overall upward/downward trends and explicitly identify the top $k$ extreme points.
\textcolor{Violet}{In addition, we further investigated the impact of revised text types on generation performance. These results support our findings in the TSF task, such as rule-based contextual information may bring more confusion, as detailed in the Appendix~\ref{texttype_diffusion}.}

\begin{table}[t]
    \centering
    \begin{adjustbox}{width=\columnwidth}
    \begin{tabular}{lcccc}
        \toprule
        \textbf{Text Types} &  \multicolumn{2}{c}{\textbf{AirPassenger}}  &  \multicolumn{2}{c}{\textbf{Sunspots}}  \\
        \cmidrule(lr){2-3} \cmidrule(lr){4-5}   
        {} & LLMTime & LSTPrompt & LLMTime & LSTPrompt \\
        \midrule
        \textit{Rule-based Text}  & $52.41$ & $20.08$ & $63.92$ & $51.61$\\
        \textit{Initial Text}  & $49.36$ & $15.12$ & $59.88$ & $49.71$\\
        \textit{Refined Text(Ours)}  & \underline{$40.94$} & \underline{$12.39$} & \underline{$48.64$} & \underline{$42.37$}\\   
        \hdashline
        \textit{w/o Instance Context}  & $41.96$ & $13.54$ & $54.33$ & $4 4.23$\\
        \textit{w/o Background}  & $44.63$ & $14.77$ & $56.81$ & $46.07$ \\
        \textit{w/o Statistical Context}  & $44.01$ & $13.41$ & $54.24$ & $47.12$ \\
        \textit{w/o Pattern}  & $44.36$ & $14.52$ & $55.16$ & $46.84$ \\
        \textit{w/o Pattern+Statistic}  & $44.30$ & $14.27$& $56.89$ & $45.65$ \\
        \hdashline
        \textit{Baseline~\citep{DBLP:conf/acl/LiuZWKP24}}  &  $45.75$ & $15.00$ & $59.91$ & $47.59$ \\
        \bottomrule
    \end{tabular}
    \end{adjustbox}
    \caption{\xc{Zero-shot time-series forecasting performance (Mean Absolute Error, lower is better) using different textual descriptions across two datasets: \textbf{AirPassenger} and \textbf{Sunspots}.}}
    \label{tab:zeroshot_testing_strategy}
\end{table}

\subsection{Performance Analysis on Individual Components of Multi-Agent System}
\label{7.7}

\textcolor{Violet}{\textbf{Objective:} We further evaluated different agent strategies on refined text. In the strategy experiments, the Macro approach involves a single team making high-level information adjustments, while the Micro approach emphasizes fine-grained details. The Multiple Teams strategy represents a collaborative setting where two teams work together to accomplish the task.}

As shown in Table~\ref{tab:zeroshot_testing_agent}, the multi-agent team strategy consistently achieves lower MAE than both micro- and macro-level single-agent approaches across datasets and text types. This suggests that collaborative generation captures a broader range of relevant patterns, leading to more effective forecasting guidance. Between single-agent strategies, macro-level descriptions outperform micro-level ones, indicating that concise, high-level summaries are easier for models to utilize than overly detailed inputs. Notably, all strategies perform better under the LSTPrompt setting, highlighting its stronger alignment with model input expectations. These findings underscore the importance of structured, semantically aligned textual inputs in enabling robust zero-shot forecasting performance.

\begin{table}[t]
    \centering
    \begin{adjustbox}{width=\columnwidth}
    \begin{tabular}{lcccc}
        \toprule
        \textbf{Text Types} &  \multicolumn{2}{c}{\textbf{AirPassenger}}  &  \multicolumn{2}{c}{\textbf{Sunspots}}  \\
        \cmidrule(lr){2-3} \cmidrule(lr){4-5}   
        {} & LLMTime & LSTPrompt & LLMTime & LSTPrompt \\
        \midrule
        \textbf{Agent Strategies} &  \multicolumn{2}{c}{}  &  \multicolumn{2}{c}{}  \\
        \textit{Multi-Agent Teams}  & $40.94$ & $12.39$ & $48.64$ & $42.37$\\ 
        \textit{Single (Micro)}  & $44.27$ & $14.22$ & $56.80$ & $45.70$\\ 
        \textit{Single (Macro)}  & $42.57$ & $13.83$ & $54.51$ & $45.01$\\ 
        \bottomrule
    \end{tabular}
    \end{adjustbox}
    \caption{Zero-shot TS forecasting performance under different multi-agent strategies. MAE is reported for each configuration; lower values indicate better performance.}
    \label{tab:zeroshot_testing_agent}
\end{table}

\subsection{Ablation Study} 
\label{7.6}
\textcolor{Violet}{\textbf{Objective:} We further conducted ablation experiments to explore the impact of different components in the multi-agent system, as well as the performance of different settings, varying prototype quantities, and using different language models as the text encoder in \frameworkname{}.}

As shown in Table~\ref{tab:mdd_kl_main}, the \frameworkname{} outperforms other variants on most datasets, as shown by its superior MDD and K-L divergence scores. Removing text input (\frameworkname{} w/o Text) leads to higher MDD and K-L divergence in nearly all datasets, highlighting the importance of text for improving generation quality. In the NN5 dataset, MDD increases from $0.591$ to $0.628$, and in the Exchange dataset, K-L divergence rises from $0.083$ to $0.088$. The removal of prototypes (\frameworkname{} w/o Prototypes) causes the most significant performance decline. As shown in Table~\ref{tab:ablation_prototypes} in Appendix~\ref{prototype}, the number of prototypes significantly influences performance: more prototypes provide richer reference patterns that enhance generation quality. However, we also observe that increasing the number of prototypes beyond 16 brings only marginal improvement. Therefore, we adopt 16 prototypes as a practical trade-off between performance and computational efficiency.
\textcolor{Violet}{Moreover, we explored the impact of pre-training knowledge from LLMs \citep{DBLP:journals/corr/abs-2407-21783}. The results show that the larger models have a slight change in performance, but it is not significant, indicating that the pre-training knowledge has a minor influence on performance (shown in Appendix~\ref{llm_impact}). For visualization of semantics on various datasets, please refer to the Appendix~\ref{prototypes_domain_sample}.}

\section{Conclusion}
In this work, we explore the potential of using text to guide TSG. We propose a multi-agent system for refining textual descriptions and a text-controlled TSG model.
Experiments show that concise text improves text-controlled performance, with our model surpassing baselines, especially in few-shot learning, demonstrating strong generalization. 
Additionally, results indicate that the designed semantic prototypes effectively complemented domain information. 
These findings lay the groundwork for advancing human-preferred text-controlled TSG.

\section*{Impact Statement}

This paper presents work aimed at advancing the field of Time-series Generation (TSG) by introducing a novel framework, \frameworkname{}, which incorporates text to guide and improve time series generation. The potential societal consequences of our work are significant, as TSG plays a crucial role in applications ranging from simulations to counterfactual analysis. By enabling controlled generation of time series tailored to domain-specific constraints and instance-level requirements, our approach has the potential to drive innovations in fields such as healthcare, finance, and climate modeling. Ethically, the datasets we use are publicly available research datasets that have already undergone ethical review.

\section*{Acknowledgement}
Viktor Schlegel is part of the IN-CYPHER programme and is supported by the National Research Foundation, Prime Minister’s Office, Singapore, under its Campus for Research Excellence and Technological Enterprise (CREATE) programme. We are especially grateful to Microsoft Research for providing the computational resources that supported this work. We also thank Research IT at the University of Manchester for access to the Computational Shared Facility, and the Imperial College Research Computing Service\footnote{DOI: \url{https://doi.org/10.14469/hpc/2232}} for additional computing support.


\bibliography{example_paper}

\begin{thebibliography}{72}
\providecommand{\natexlab}[1]{#1}
\providecommand{\url}[1]{\texttt{#1}}
\expandafter\ifx\csname urlstyle\endcsname\relax
  \providecommand{\doi}[1]{doi: #1}\else
  \providecommand{\doi}{doi: \begingroup \urlstyle{rm}\Url}\fi

\bibitem[Alcaraz \& Strodthoff(2023)Alcaraz and Strodthoff]{DBLP:journals/tmlr/AlcarazS23}
Alcaraz, J. M.~L. and Strodthoff, N.
\newblock Diffusion-based time series imputation and forecasting with structured state space models.
\newblock \emph{Trans. Mach. Learn. Res.}, 2023, 2023.

\bibitem[Alexandrov et~al.(2020)Alexandrov, Benidis, Bohlke-Schneider, Flunkert, Gasthaus, Januschowski, Maddix, Rangapuram, Salinas, Schulz, Stella, Türkmen, and Wang]{gluonts_jmlr}
Alexandrov, A., Benidis, K., Bohlke-Schneider, M., Flunkert, V., Gasthaus, J., Januschowski, T., Maddix, D.~C., Rangapuram, S., Salinas, D., Schulz, J., Stella, L., Türkmen, A.~C., and Wang, Y.
\newblock {GluonTS: Probabilistic and Neural Time Series Modeling in Python}.
\newblock \emph{Journal of Machine Learning Research}, 21\penalty0 (116):\penalty0 1--6, 2020.
\newblock URL \url{http://jmlr.org/papers/v21/19-820.html}.

\bibitem[Ansari et~al.(2024)Ansari, Stella, T{\"{u}}rkmen, Zhang, Mercado, Shen, Shchur, Rangapuram, Pineda{-}Arango, Kapoor, Zschiegner, Maddix, Mahoney, Torkkola, Wilson, Bohlke{-}Schneider, and Wang]{DBLP:journals/corr/abs-2403-07815}
Ansari, A.~F., Stella, L., T{\"{u}}rkmen, A.~C., Zhang, X., Mercado, P., Shen, H., Shchur, O., Rangapuram, S.~S., Pineda{-}Arango, S., Kapoor, S., Zschiegner, J., Maddix, D.~C., Mahoney, M.~W., Torkkola, K., Wilson, A.~G., Bohlke{-}Schneider, M., and Wang, Y.
\newblock Chronos: Learning the language of time series.
\newblock \emph{CoRR}, abs/2403.07815, 2024.

\bibitem[Bao et~al.(2024)Bao, Ang, Huang, Tung, and Huang]{DBLP:journals/corr/abs-2403-03698}
Bao, Y., Ang, Y., Huang, Q., Tung, A. K.~H., and Huang, Z.
\newblock Towards controllable time series generation.
\newblock \emph{CoRR}, abs/2403.03698, 2024.

\bibitem[Berger \& Zhou(2014)Berger and Zhou]{berger2014kolmogorov}
Berger, V.~W. and Zhou, Y.
\newblock Kolmogorov--smirnov test: Overview.
\newblock \emph{Wiley statsref: Statistics reference online}, 2014.

\bibitem[Cao et~al.(2024)Cao, Jia, Arik, Pfister, Zheng, Ye, and Liu]{DBLP:conf/iclr/CaoJAPZY024}
Cao, D., Jia, F., Arik, S.~{\"{O}}., Pfister, T., Zheng, Y., Ye, W., and Liu, Y.
\newblock {TEMPO:} prompt-based generative pre-trained transformer for time series forecasting.
\newblock In \emph{{ICLR}}. OpenReview.net, 2024.

\bibitem[Chang et~al.(2023)Chang, Peng, and Chen]{DBLP:journals/corr/abs-2308-08469}
Chang, C., Peng, W., and Chen, T.
\newblock {LLM4TS:} two-stage fine-tuning for time-series forecasting with pre-trained llms.
\newblock \emph{CoRR}, abs/2308.08469, 2023.

\bibitem[Chen et~al.(2023)Chen, Zheng, Lu, Yuan, and Zhu]{DBLP:journals/corr/abs-2306-03763}
Chen, Z., Zheng, L.~N., Lu, C., Yuan, J., and Zhu, D.
\newblock Chatgpt informed graph neural network for stock movement prediction.
\newblock \emph{CoRR}, abs/2306.03763, 2023.

\bibitem[Cleveland et~al.(1990)Cleveland, Cleveland, McRae, Terpenning, et~al.]{cleveland1990stl}
Cleveland, R.~B., Cleveland, W.~S., McRae, J.~E., Terpenning, I., et~al.
\newblock Stl: A seasonal-trend decomposition.
\newblock \emph{J. off. Stat}, 6\penalty0 (1):\penalty0 3--73, 1990.

\bibitem[Coletta et~al.(2023)Coletta, Gopalakrishnan, Borrajo, and Vyetrenko]{DBLP:conf/nips/ColettaGBV23}
Coletta, A., Gopalakrishnan, S., Borrajo, D., and Vyetrenko, S.
\newblock On the constrained time-series generation problem.
\newblock In \emph{NeurIPS}, 2023.

\bibitem[Deng et~al.(2025)Deng, Xu, Li, Huang, Hou, and Bian]{deng2025tardiff}
Deng, B., Xu, C., Li, H., Huang, Y., Hou, M., and Bian, J.
\newblock Tardiff: Target-oriented diffusion guidance for synthetic electronic health record time series generation.
\newblock \emph{arXiv preprint arXiv:2504.17613}, 2025.

\bibitem[Desai et~al.(2021)Desai, Freeman, Wang, and Beaver]{DBLP:journals/corr/abs-2111-08095}
Desai, A., Freeman, C., Wang, Z., and Beaver, I.
\newblock Timevae: {A} variational auto-encoder for multivariate time series generation.
\newblock \emph{CoRR}, abs/2111.08095, 2021.

\bibitem[Dubey et~al.(2024)Dubey, Jauhri, Pandey, Kadian, Al{-}Dahle, Letman, Mathur, Schelten, Yang, Fan, Goyal, Hartshorn, Yang, Mitra, Sravankumar, Korenev, Hinsvark, Rao, Zhang, Rodriguez, Gregerson, Spataru, Rozi{\`{e}}re, Biron, Tang, Chern, Caucheteux, Nayak, Bi, Marra, McConnell, Keller, Touret, Wu, Wong, Ferrer, Nikolaidis, Allonsius, Song, Pintz, Livshits, Esiobu, Choudhary, Mahajan, Garcia{-}Olano, Perino, Hupkes, Lakomkin, AlBadawy, Lobanova, Dinan, Smith, Radenovic, Zhang, Synnaeve, Lee, Anderson, Nail, Mialon, Pang, Cucurell, Nguyen, Korevaar, Xu, Touvron, Zarov, Ibarra, Kloumann, Misra, Evtimov, Copet, Lee, Geffert, Vranes, Park, Mahadeokar, Shah, van~der Linde, Billock, Hong, Lee, Fu, Chi, Huang, Liu, Wang, Yu, Bitton, Spisak, Park, Rocca, Johnstun, Saxe, Jia, Alwala, Upasani, Plawiak, Li, Heafield, Stone, and et~al.]{DBLP:journals/corr/abs-2407-21783}
Dubey, A., Jauhri, A., Pandey, A., Kadian, A., Al{-}Dahle, A., Letman, A., Mathur, A., Schelten, A., Yang, A., Fan, A., Goyal, A., Hartshorn, A., Yang, A., Mitra, A., Sravankumar, A., Korenev, A., Hinsvark, A., Rao, A., Zhang, A., Rodriguez, A., Gregerson, A., Spataru, A., Rozi{\`{e}}re, B., Biron, B., Tang, B., Chern, B., Caucheteux, C., Nayak, C., Bi, C., Marra, C., McConnell, C., Keller, C., Touret, C., Wu, C., Wong, C., Ferrer, C.~C., Nikolaidis, C., Allonsius, D., Song, D., Pintz, D., Livshits, D., Esiobu, D., Choudhary, D., Mahajan, D., Garcia{-}Olano, D., Perino, D., Hupkes, D., Lakomkin, E., AlBadawy, E., Lobanova, E., Dinan, E., Smith, E.~M., Radenovic, F., Zhang, F., Synnaeve, G., Lee, G., Anderson, G.~L., Nail, G., Mialon, G., Pang, G., Cucurell, G., Nguyen, H., Korevaar, H., Xu, H., Touvron, H., Zarov, I., Ibarra, I.~A., Kloumann, I.~M., Misra, I., Evtimov, I., Copet, J., Lee, J., Geffert, J., Vranes, J., Park, J., Mahadeokar, J., Shah, J., van~der Linde, J., Billock, J., Hong, J., Lee, J., Fu,
  J., Chi, J., Huang, J., Liu, J., Wang, J., Yu, J., Bitton, J., Spisak, J., Park, J., Rocca, J., Johnstun, J., Saxe, J., Jia, J., Alwala, K.~V., Upasani, K., Plawiak, K., Li, K., Heafield, K., Stone, K., and et~al.
\newblock The llama 3 herd of models.
\newblock \emph{CoRR}, abs/2407.21783, 2024.

\bibitem[Fan et~al.(2024)Fan, Wu, Xu, Huang, Liu, and Bian]{fan2024mg}
Fan, X., Wu, Y., Xu, C., Huang, Y., Liu, W., and Bian, J.
\newblock Mg-tsd: Multi-granularity time series diffusion models with guided learning process.
\newblock \emph{arXiv preprint arXiv:2403.05751}, 2024.

\bibitem[Godahewa et~al.(2021)Godahewa, Bergmeir, Webb, Hyndman, and Montero-Manso]{godahewa2021monash}
Godahewa, R., Bergmeir, C., Webb, G.~I., Hyndman, R.~J., and Montero-Manso, P.
\newblock Monash time series forecasting archive.
\newblock In \emph{Neural Information Processing Systems Track on Datasets and Benchmarks}, 2021.

\bibitem[Gruver et~al.(2023)Gruver, Finzi, Qiu, and Wilson]{DBLP:conf/nips/GruverFQW23}
Gruver, N., Finzi, M., Qiu, S., and Wilson, A.~G.
\newblock Large language models are zero-shot time series forecasters.
\newblock In \emph{NeurIPS}, 2023.

\bibitem[Gunjal \& Durrett(2023)Gunjal and Durrett]{DBLP:journals/corr/abs-2305-14847}
Gunjal, A. and Durrett, G.
\newblock Drafting event schemas using language models.
\newblock \emph{CoRR}, abs/2305.14847, 2023.

\bibitem[Guo et~al.(2024)Guo, Wang, Guo, Li, Song, Tan, Liu, Bian, and Yang]{DBLP:conf/iclr/Guo0GLS0L0Y24}
Guo, Q., Wang, R., Guo, J., Li, B., Song, K., Tan, X., Liu, G., Bian, J., and Yang, Y.
\newblock Connecting large language models with evolutionary algorithms yields powerful prompt optimizers.
\newblock In \emph{{ICLR}}. OpenReview.net, 2024.

\bibitem[Harris \& Zaki(2022)Harris and Zaki]{DBLP:journals/corr/abs-2207-05194}
Harris, J.~J. and Zaki, M.~J.
\newblock Towards neural numeric-to-text generation from temporal personal health data.
\newblock \emph{CoRR}, abs/2207.05194, 2022.
\newblock \doi{10.48550/ARXIV.2207.05194}.
\newblock URL \url{https://doi.org/10.48550/arXiv.2207.05194}.

\bibitem[Hasnain et~al.(2022)Hasnain, Sheng, Hashmi, Bhatti, Hussain, Hameed, Marjan, Bazai, Hossain, Sahabuddin, et~al.]{hasnain2022time}
Hasnain, A., Sheng, Y., Hashmi, M.~Z., Bhatti, U.~A., Hussain, A., Hameed, M., Marjan, S., Bazai, S.~U., Hossain, M.~A., Sahabuddin, M., et~al.
\newblock Time series analysis and forecasting of air pollutants based on prophet forecasting model in jiangsu province, china.
\newblock \emph{Frontiers in Environmental Science}, 10:\penalty0 945628, 2022.

\bibitem[Ho et~al.(2020)Ho, Jain, and Abbeel]{ho2020denoising}
Ho, J., Jain, A., and Abbeel, P.
\newblock Denoising diffusion probabilistic models.
\newblock \emph{Advances in neural information processing systems}, 33:\penalty0 6840--6851, 2020.

\bibitem[Hong \& Chun(2023)Hong and Chun]{DBLP:journals/bspc/HongC23}
Hong, J. and Chun, H.
\newblock A prediction model for healthcare time-series data with a mixture of deep mixed effect models using gaussian processes.
\newblock \emph{Biomed. Signal Process. Control.}, 84:\penalty0 104753, 2023.

\bibitem[Hou et~al.(2024)Hou, Wu, Xu, Huang, Bai, Wu, and Bian]{hou2024invdiff}
Hou, M., Wu, Y., Xu, C., Huang, Y.-H., Bai, C., Wu, L., and Bian, J.
\newblock Invdiff: Invariant guidance for bias mitigation in diffusion models.
\newblock \emph{arXiv preprint arXiv:2412.08480}, 2024.

\bibitem[Huang \& Deng(2023)Huang and Deng]{DBLP:journals/nn/HuangD23}
Huang, F. and Deng, Y.
\newblock {TCGAN:} convolutional generative adversarial network for time series classification and clustering.
\newblock \emph{Neural Networks}, 165:\penalty0 868--883, 2023.

\bibitem[Huang et~al.(2024)Huang, Xu, Liu, Liu, Li, and Bian]{DBLP:journals/corr/abs-2408-12991}
Huang, Y., Xu, C., Liu, Y., Liu, W., Li, W., and Bian, J.
\newblock Controllable financial market generation with diffusion guided meta agent.
\newblock \emph{CoRR}, abs/2408.12991, 2024.

\bibitem[Huang et~al.(2025)Huang, Xu, Wu, Li, and Bian]{huang2025timedp}
Huang, Y.-H., Xu, C., Wu, Y., Li, W.-J., and Bian, J.
\newblock Timedp: Learning to generate multi-domain time series with domain prompts.
\newblock \emph{arXiv preprint arXiv:2501.05403}, 2025.

\bibitem[Hurvich(1988)]{hurvich1988mean}
Hurvich, C.~M.
\newblock A mean squared error criterion for time series data windows.
\newblock \emph{Biometrika}, 75\penalty0 (3):\penalty0 485--490, 1988.

\bibitem[Jeon et~al.(2022)Jeon, Kim, Song, Cho, and Park]{DBLP:conf/nips/JeonKSCP22}
Jeon, J., Kim, J., Song, H., Cho, S., and Park, N.
\newblock {GT-GAN:} general purpose time series synthesis with generative adversarial networks.
\newblock In \emph{NeurIPS}, 2022.

\bibitem[Jin et~al.(2023)Jin, Wang, Ma, Chu, Zhang, Shi, Chen, Liang, Li, Pan, and Wen]{DBLP:journals/corr/abs-2310-01728}
Jin, M., Wang, S., Ma, L., Chu, Z., Zhang, J.~Y., Shi, X., Chen, P., Liang, Y., Li, Y., Pan, S., and Wen, Q.
\newblock Time-llm: Time series forecasting by reprogramming large language models.
\newblock \emph{CoRR}, abs/2310.01728, 2023.

\bibitem[Kollovieh et~al.(2024)Kollovieh, Ansari, Bohlke-Schneider, Zschiegner, Wang, and Wang]{kollovieh2024predict}
Kollovieh, M., Ansari, A.~F., Bohlke-Schneider, M., Zschiegner, J., Wang, H., and Wang, Y.~B.
\newblock Predict, refine, synthesize: Self-guiding diffusion models for probabilistic time series forecasting.
\newblock \emph{Advances in Neural Information Processing Systems}, 36, 2024.

\bibitem[Lee et~al.(2023)Lee, Malacarne, and Aune]{DBLP:conf/aistats/LeeMA23}
Lee, D., Malacarne, S., and Aune, E.
\newblock Vector quantized time series generation with a bidirectional prior model.
\newblock In \emph{{AISTATS}}, volume 206 of \emph{Proceedings of Machine Learning Research}, pp.\  7665--7693. {PMLR}, 2023.

\bibitem[Li et~al.(2023)Li, Wu, Schlegel, Batista{-}Navarro, Nguyen, Kashyap, Zeng, Beck, Winkler, and Nenadic]{DBLP:conf/bionlp/LiWSBNKZB0N23}
Li, H., Wu, Y., Schlegel, V., Batista{-}Navarro, R., Nguyen, T., Kashyap, A.~R., Zeng, X., Beck, D., Winkler, S., and Nenadic, G.
\newblock Team: {PULSAR} at probsum 2023: {PULSAR:} pre-training with extracted healthcare terms for summarising patients' problems and data augmentation with black-box large language models.
\newblock In \emph{BioNLP@ACL}, pp.\  503--509. Association for Computational Linguistics, 2023.

\bibitem[Liang et~al.(2024)Liang, Wen, Nie, Jiang, Jin, Song, Pan, and Wen]{DBLP:conf/kdd/LiangWNJ0SPW24}
Liang, Y., Wen, H., Nie, Y., Jiang, Y., Jin, M., Song, D., Pan, S., and Wen, Q.
\newblock Foundation models for time series analysis: {A} tutorial and survey.
\newblock In \emph{{KDD}}, pp.\  6555--6565. {ACM}, 2024.

\bibitem[Lin et~al.(2024)Lin, Li, Li, Li, and Gao]{lin2024diffusion}
Lin, L., Li, Z., Li, R., Li, X., and Gao, J.
\newblock Diffusion models for time-series applications: a survey.
\newblock \emph{Frontiers of Information Technology \& Electronic Engineering}, 25\penalty0 (1):\penalty0 19--41, 2024.

\bibitem[Liu et~al.(2024{\natexlab{a}})Liu, Xu, Zhao, Kong, Kamarthi, Sasanur, Sharma, Cui, Wen, Zhang, and Prakash]{DBLP:conf/nips/LiuXZKKSSCW0P24}
Liu, H., Xu, S., Zhao, Z., Kong, L., Kamarthi, H., Sasanur, A.~B., Sharma, M., Cui, J., Wen, Q., Zhang, C., and Prakash, B.~A.
\newblock Time-mmd: Multi-domain multimodal dataset for time series analysis.
\newblock In \emph{NeurIPS}, 2024{\natexlab{a}}.

\bibitem[Liu et~al.(2024{\natexlab{b}})Liu, Zhao, Wang, Kamarthi, and Prakash]{DBLP:conf/acl/LiuZWKP24}
Liu, H., Zhao, Z., Wang, J., Kamarthi, H., and Prakash, B.~A.
\newblock Lstprompt: Large language models as zero-shot time series forecasters by long-short-term prompting.
\newblock In \emph{{ACL} (Findings)}, pp.\  7832--7840. Association for Computational Linguistics, 2024{\natexlab{b}}.

\bibitem[Liu et~al.(2024{\natexlab{c}})Liu, Chen, Qu, Tang, and Ong]{DBLP:conf/cec/LiuCQ0O24}
Liu, S., Chen, C., Qu, X., Tang, K., and Ong, Y.
\newblock Large language models as evolutionary optimizers.
\newblock In \emph{{CEC}}, pp.\  1--8. {IEEE}, 2024{\natexlab{c}}.

\bibitem[Liu et~al.(2024{\natexlab{d}})Liu, Hu, Li, Diao, Liang, Hooi, and Zimmermann]{liu2024unitime}
Liu, X., Hu, J., Li, Y., Diao, S., Liang, Y., Hooi, B., and Zimmermann, R.
\newblock Unitime: A language-empowered unified model for cross-domain time series forecasting.
\newblock In \emph{Proceedings of the ACM on Web Conference 2024}, pp.\  4095--4106, 2024{\natexlab{d}}.

\bibitem[Liu et~al.(2024{\natexlab{e}})Liu, Zhang, Li, Yan, Gao, Chen, Yuan, Huang, Sun, Gao, He, and Sun]{DBLP:journals/corr/abs-2402-17177}
Liu, Y., Zhang, K., Li, Y., Yan, Z., Gao, C., Chen, R., Yuan, Z., Huang, Y., Sun, H., Gao, J., He, L., and Sun, L.
\newblock Sora: {A} review on background, technology, limitations, and opportunities of large vision models.
\newblock \emph{CoRR}, abs/2402.17177, 2024{\natexlab{e}}.

\bibitem[Lopez{-}Lira \& Tang(2023)Lopez{-}Lira and Tang]{DBLP:journals/corr/abs-2304-07619}
Lopez{-}Lira, A. and Tang, Y.
\newblock Can chatgpt forecast stock price movements? return predictability and large language models.
\newblock \emph{CoRR}, abs/2304.07619, 2023.

\bibitem[Madaan et~al.(2023)Madaan, Tandon, Gupta, Hallinan, Gao, Wiegreffe, Alon, Dziri, Prabhumoye, Yang, Gupta, Majumder, Hermann, Welleck, Yazdanbakhsh, and Clark]{DBLP:conf/nips/MadaanTGHGW0DPY23}
Madaan, A., Tandon, N., Gupta, P., Hallinan, S., Gao, L., Wiegreffe, S., Alon, U., Dziri, N., Prabhumoye, S., Yang, Y., Gupta, S., Majumder, B.~P., Hermann, K., Welleck, S., Yazdanbakhsh, A., and Clark, P.
\newblock Self-refine: Iterative refinement with self-feedback.
\newblock In \emph{NeurIPS}, 2023.

\bibitem[Makridakis et~al.(2018)Makridakis, Spiliotis, and Assimakopoulos]{makridakis2018m4}
Makridakis, S., Spiliotis, E., and Assimakopoulos, V.
\newblock The m4 competition: Results, findings, conclusion and way forward.
\newblock \emph{International Journal of forecasting}, 34\penalty0 (4):\penalty0 802--808, 2018.

\bibitem[Merrill et~al.(2024)Merrill, Tan, Gupta, Hartvigsen, and Althoff]{DBLP:journals/corr/abs-2404-11757}
Merrill, M.~A., Tan, M., Gupta, V., Hartvigsen, T., and Althoff, T.
\newblock Language models still struggle to zero-shot reason about time series.
\newblock \emph{CoRR}, abs/2404.11757, 2024.

\bibitem[Narasimhan et~al.(2024)Narasimhan, Agarwal, Akcin, Sanghavi, and Chinchali]{DBLP:conf/icml/NarasimhanAASC24}
Narasimhan, S.~S., Agarwal, S., Akcin, O., Sanghavi, S., and Chinchali, S.~P.
\newblock Time weaver: {A} conditional time series generation model.
\newblock In \emph{{ICML}}. OpenReview.net, 2024.

\bibitem[Nie et~al.(2023)Nie, Nguyen, Sinthong, and Kalagnanam]{patchtst}
Nie, Y., Nguyen, N.~H., Sinthong, P., and Kalagnanam, J.
\newblock A time series is worth 64 words: Long-term forecasting with transformers.
\newblock In \emph{International Conference on Learning Representations}, 2023.

\bibitem[Oreshkin et~al.(2020)Oreshkin, Carpov, Chapados, and Bengio]{DBLP:conf/iclr/OreshkinCCB20}
Oreshkin, B.~N., Carpov, D., Chapados, N., and Bengio, Y.
\newblock {N-BEATS:} neural basis expansion analysis for interpretable time series forecasting.
\newblock In \emph{{ICLR}}. OpenReview.net, 2020.

\bibitem[Panaretos \& Zemel(2019)Panaretos and Zemel]{panaretos2019statistical}
Panaretos, V.~M. and Zemel, Y.
\newblock Statistical aspects of wasserstein distances.
\newblock \emph{Annual review of statistics and its application}, 6:\penalty0 405--431, 2019.

\bibitem[P{\"o}hl et~al.(2025)P{\"o}hl, Schlegel, Li, and Bharath]{pohl2025generating}
P{\"o}hl, P., Schlegel, V., Li, H., and Bharath, A.
\newblock Generating realistic multi-beat ecg signals.
\newblock \emph{arXiv preprint arXiv:2505.18189}, 2025.

\bibitem[Qian et~al.(2024)Qian, Xie, Wan, Li, Sun, and Chiang]{qian2024timeldm}
Qian, J., Xie, B., Wan, B., Li, M., Sun, M., and Chiang, P.~Y.
\newblock Timeldm: Latent diffusion model for unconditional time series generation.
\newblock \emph{arXiv preprint arXiv:2407.04211}, 2024.

\bibitem[Rasul et~al.(2021)Rasul, Seward, Schuster, and Vollgraf]{rasul2021autoregressive}
Rasul, K., Seward, C., Schuster, I., and Vollgraf, R.
\newblock Autoregressive denoising diffusion models for multivariate probabilistic time series forecasting.
\newblock In \emph{International Conference on Machine Learning}, pp.\  8857--8868. PMLR, 2021.

\bibitem[Schlegel et~al.(2023)Schlegel, Li, Wu, Subramanian, Nguyen, Kashyap, Beck, Zeng, Batista{-}Navarro, Winkler, and Nenadic]{DBLP:conf/clef/SchlegelLW0NKBZ23}
Schlegel, V., Li, H., Wu, Y., Subramanian, A., Nguyen, T., Kashyap, A.~R., Beck, D., Zeng, X., Batista{-}Navarro, R.~T., Winkler, S., and Nenadic, G.
\newblock {PULSAR} at mediqa-sum 2023: Large language models augmented by synthetic dialogue convert patient dialogues to medical records.
\newblock In \emph{{CLEF} (Working Notes)}, volume 3497 of \emph{{CEUR} Workshop Proceedings}, pp.\  1668--1679. CEUR-WS.org, 2023.

\bibitem[Sezer et~al.(2020)Sezer, Gudelek, and {\"{O}}zbayoglu]{DBLP:journals/asc/SezerGO20}
Sezer, O.~B., Gudelek, M.~U., and {\"{O}}zbayoglu, A.~M.
\newblock Financial time series forecasting with deep learning : {A} systematic literature review: 2005-2019.
\newblock \emph{Appl. Soft Comput.}, 90:\penalty0 106181, 2020.

\bibitem[Shen \& Kwok(2023)Shen and Kwok]{DBLP:conf/icml/ShenK23}
Shen, L. and Kwok, J.~T.
\newblock Non-autoregressive conditional diffusion models for time series prediction.
\newblock In \emph{{ICML}}, volume 202 of \emph{Proceedings of Machine Learning Research}, pp.\  31016--31029. {PMLR}, 2023.

\bibitem[Shen et~al.(2024)Shen, Chen, and Kwok]{DBLP:conf/iclr/ShenCK24}
Shen, L., Chen, W., and Kwok, J.~T.
\newblock Multi-resolution diffusion models for time series forecasting.
\newblock In \emph{{ICLR}}. OpenReview.net, 2024.

\bibitem[Shi et~al.(2024)Shi, Xue, Wang, Zhou, Zhang, Zhou, Tan, and Mei]{shi2024language}
Shi, X., Xue, S., Wang, K., Zhou, F., Zhang, J., Zhou, J., Tan, C., and Mei, H.
\newblock Language models can improve event prediction by few-shot abductive reasoning.
\newblock \emph{Advances in Neural Information Processing Systems}, 36, 2024.

\bibitem[T et~al.(2024)T, Agarwal, Sanjay, Sarda, Alex, Gupta, Kumar, and Kamath]{DBLP:conf/spcom/TASSAGKK24}
T, K.~J., Agarwal, A., Sanjay, S., Sarda, Y., Alex, J. S.~R., Gupta, S., Kumar, S., and Kamath, V.
\newblock Thread detection and response generation using transformers with prompt optimisation.
\newblock In \emph{{SPCOM}}, pp.\  1--5. {IEEE}, 2024.

\bibitem[Tashiro et~al.(2021)Tashiro, Song, Song, and Ermon]{DBLP:conf/nips/TashiroSSE21}
Tashiro, Y., Song, J., Song, Y., and Ermon, S.
\newblock {CSDI:} conditional score-based diffusion models for probabilistic time series imputation.
\newblock In \emph{NeurIPS}, pp.\  24804--24816, 2021.

\bibitem[Wei et~al.(2022)Wei, Wang, Schuurmans, Bosma, Ichter, Xia, Chi, Le, and Zhou]{DBLP:conf/nips/Wei0SBIXCLZ22}
Wei, J., Wang, X., Schuurmans, D., Bosma, M., Ichter, B., Xia, F., Chi, E.~H., Le, Q.~V., and Zhou, D.
\newblock Chain-of-thought prompting elicits reasoning in large language models.
\newblock In \emph{NeurIPS}, 2022.

\bibitem[Westgaard et~al.(2021)Westgaard, Fleten, Negash, Botterud, Bogaard, and Verling]{westgaard2021performing}
Westgaard, S., Fleten, S.-E., Negash, A., Botterud, A., Bogaard, K., and Verling, T.~H.
\newblock Performing price scenario analysis and stress testing using quantile regression: A case study of the californian electricity market.
\newblock \emph{Energy}, 214:\penalty0 118796, 2021.

\bibitem[Woo et~al.(2024)Woo, Liu, Kumar, Xiong, Savarese, and Sahoo]{moirai}
Woo, G., Liu, C., Kumar, A., Xiong, C., Savarese, S., and Sahoo, D.
\newblock Unified training of universal time series forecasting transformers.
\newblock In \emph{Proceedings of the International Conference on Machine Learning}, pp.\  53140--53164, 2024.

\bibitem[Wu et~al.(2023)Wu, Hu, Liu, Zhou, Wang, and Long]{DBLP:conf/iclr/WuHLZ0L23}
Wu, H., Hu, T., Liu, Y., Zhou, H., Wang, J., and Long, M.
\newblock Timesnet: Temporal 2d-variation modeling for general time series analysis.
\newblock In \emph{{ICLR}}. OpenReview.net, 2023.

\bibitem[Xie et~al.(2023)Xie, Han, Lai, Peng, and Huang]{xie2023wall}
Xie, Q., Han, W., Lai, Y., Peng, M., and Huang, J.
\newblock The wall street neophyte: A zero-shot analysis of chatgpt over multimodal stock movement prediction challenges.
\newblock \emph{arXiv preprint arXiv:2304.05351}, 2023.

\bibitem[Xue \& Salim(2024)Xue and Salim]{DBLP:journals/tkde/XueS24}
Xue, H. and Salim, F.~D.
\newblock Promptcast: {A} new prompt-based learning paradigm for time series forecasting.
\newblock \emph{{IEEE} Trans. Knowl. Data Eng.}, 36\penalty0 (11):\penalty0 6851--6864, 2024.

\bibitem[Yao et~al.(2023)Yao, Zhao, Yu, Du, Shafran, Narasimhan, and Cao]{DBLP:conf/iclr/YaoZYDSN023}
Yao, S., Zhao, J., Yu, D., Du, N., Shafran, I., Narasimhan, K.~R., and Cao, Y.
\newblock React: Synergizing reasoning and acting in language models.
\newblock In \emph{{ICLR}}. OpenReview.net, 2023.

\bibitem[Yoon et~al.(2019)Yoon, Jarrett, and van~der Schaar]{DBLP:conf/nips/YoonJS19}
Yoon, J., Jarrett, D., and van~der Schaar, M.
\newblock Time-series generative adversarial networks.
\newblock In \emph{NeurIPS}, pp.\  5509--5519, 2019.

\bibitem[Yuan \& Qiao(2024)Yuan and Qiao]{DBLP:conf/iclr/YuanQ24}
Yuan, X. and Qiao, Y.
\newblock Diffusion-ts: Interpretable diffusion for general time series generation.
\newblock In \emph{{ICLR}}. OpenReview.net, 2024.

\bibitem[Zhang et~al.(2024{\natexlab{a}})Zhang, Zhou, and Li]{zhang2024can}
Zhang, B., Zhou, Y., and Li, D.
\newblock Can human reading validate a topic model?
\newblock \emph{Sociological Methodology}, pp.\  00811750241265336, 2024{\natexlab{a}}.

\bibitem[Zhang et~al.(2024{\natexlab{b}})Zhang, Xu, Zhang, Zhang, Wang, Bian, and Tan]{zhang2024timeraf}
Zhang, H., Xu, C., Zhang, Y.-F., Zhang, Z., Wang, L., Bian, J., and Tan, T.
\newblock Timeraf: Retrieval-augmented foundation model for zero-shot time series forecasting.
\newblock \emph{arXiv preprint arXiv:2412.20810}, 2024{\natexlab{b}}.

\bibitem[Zheng et~al.(2023)Zheng, He, and Wang]{DBLP:journals/corr/abs-2310-02239}
Zheng, K., He, X., and Wang, X.~E.
\newblock Minigpt-5: Interleaved vision-and-language generation via generative vokens.
\newblock \emph{CoRR}, abs/2310.02239, 2023.

\bibitem[Zhou et~al.(2023{\natexlab{a}})Zhou, Niu, Wang, Sun, and Jin]{DBLP:conf/nips/ZhouNW0023}
Zhou, T., Niu, P., Wang, X., Sun, L., and Jin, R.
\newblock One fits all: Power general time series analysis by pretrained {LM}.
\newblock In \emph{NeurIPS}, 2023{\natexlab{a}}.

\bibitem[Zhou et~al.(2024)Zhou, Jia, Hu, Xie, Huang, and Yu]{DBLP:conf/infocom/ZhouJHX0Y24}
Zhou, X., Jia, Q., Hu, Y., Xie, R., Huang, T., and Yu, F.~R.
\newblock Geng: An llm-based generic time series data generation approach for edge intelligence via cross-domain collaboration.
\newblock In \emph{{INFOCOM} (Workshops)}, pp.\  1--6. {IEEE}, 2024.

\bibitem[Zhou et~al.(2023{\natexlab{b}})Zhou, Muresanu, Han, Paster, Pitis, Chan, and Ba]{DBLP:conf/iclr/ZhouMHPPCB23}
Zhou, Y., Muresanu, A.~I., Han, Z., Paster, K., Pitis, S., Chan, H., and Ba, J.
\newblock Large language models are human-level prompt engineers.
\newblock In \emph{{ICLR}}. OpenReview.net, 2023{\natexlab{b}}.

\end{thebibliography}
\bibliographystyle{icml2025}

\newpage
\appendix
\onecolumn
\section{Text Preparation}
\subsection{Multi-Agent Collaboration Framework Details}
\label{appendix_system_pipeline}

\begin{figure}[t]
    \centering
    \includegraphics[scale=1]{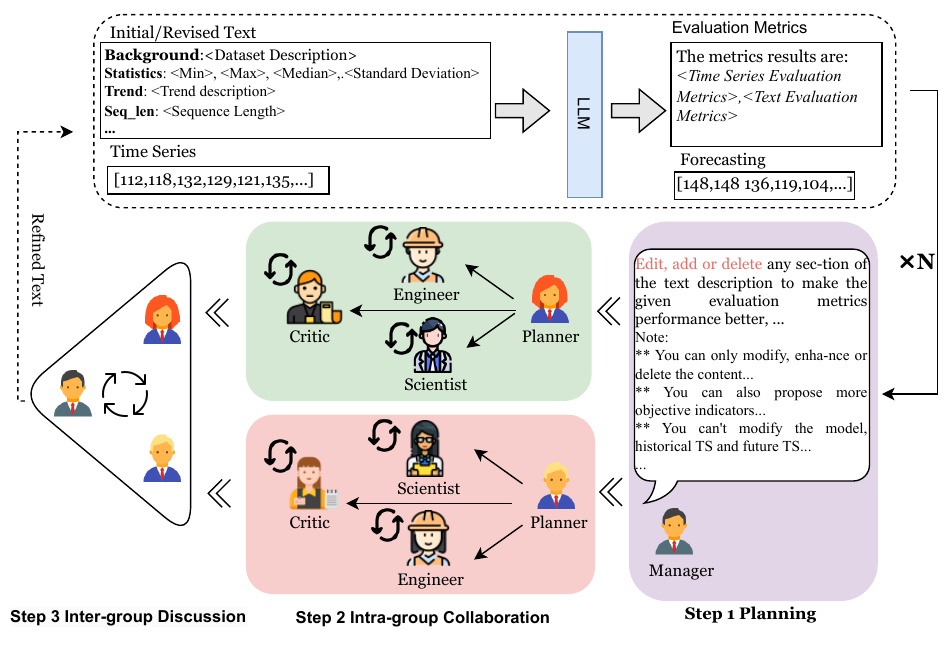}
    \caption{Detail workflow of proposed multi-agent collaborative framework}
    \label{fig:multi_agent_framework}
\end{figure}

\subsubsection{Framework Workflow}
We propose a structured, multi-agent collaboration framework designed to iteratively optimize text generation through systematic refinement. While the system is capable of operating with a single team employing distinct strategies, our experimental results demonstrate that employing two independent teams yields superior outcomes in terms of both quality and diversity of generated outputs. As can be seen from Figure  \ref{fig:multi_agent_framework}, the framework comprises three primary stages: 

In Stage 1: Task Planning, a manager agent assumes responsibility for overseeing the workflow. This agent coordinates all subsequent activities by distributing tasks and results from prior iterations to ensure seamless progress and alignment among team members. The manager also defines the objectives for the teams, thereby establishing a structured foundation for collaboration. Stage 2: Intra-group Collaboration constitutes the core of the system, wherein two independent teams of agents work concurrently to refine the given text. Each team is composed of four roles: a planner, a scientist, an engineer, and an observer. The planner serves as the team leader, formulating strategies and supervising operations. The scientist analyzes the input data and formulates detailed optimization plans. The engineer executes these plans, generating improved text outputs. The observer critically evaluates the plans and outputs, raising questions to identify shortcomings and potential improvements. Teams operate in iterative cycles, guided by the observer’s critiques. This self-refining loop continues until the observer ceases to raise objections or a predefined maximum number of iterations is reached. Through this iterative process, each team independently produces a refined output. In Stage 3: Inter-group Discussion, the leaders of the two teams engage in a structured dialogue moderated by the manager. This stage facilitates the integration of insights from both teams, encouraging comparative evaluation and collaborative refinement of their outputs. The discussion continues until a consensus is reached, resulting in a unified solution that incorporates the strengths of both teams.

The finalized output is then subjected to Post-Processing. This phase includes a validation step, where the text is evaluated against a predefined model to ensure its quality and adherence to target metrics. Approved outputs are incorporated into a formal dataset, expanding the training resources available for future tasks. Additionally, any templates developed during the process are added to a general template library, enabling reusability and continuous improvement in subsequent data generation efforts.

\subsubsection{Stages of Collaboration}

\textbf{Stage 1: Task Planning} \\
\textbf{Role Focus}: Manager  
\begin{itemize}
    \item The manager initiates the workflow and assigns responsibilities to the team leaders.
\end{itemize}

\textbf{Stage 2: Intra-group Collaboration}\\
\textbf{Role Focus}: Planner, Scientist, Engineer, and Observer  
\begin{itemize}
    \item Teams execute their tasks, with internal cycles involving:
    \begin{itemize}
        \item The scientist proposing plans.
        \item The engineer implementing them.
        \item The observer providing feedback until quality is satisfactory.
    \end{itemize}
\end{itemize}

\textbf{Stage 3: Inter-group Discussion}\\
\textbf{Role Focus}: Manager and Team Leaders  
\begin{itemize}
    \item Leaders from each team present their refined outputs to the manager.
    \item Discussions between leaders and the manager lead to a unified, optimized output.
\end{itemize}

\subsection{Example of Initial and refined text}
\label{text_dimensions}

\textbf{Initial Text:} This task focuses on predicting future values of a given time series based on historical data. The historical data shows periodic spikes in values, followed by declines, indicating a strong seasonal pattern. Notable fluctuations are observed at time points when values exceed 400, suggesting external influences. The average value of the historical data is approximately 239.95, with a variance of 8271.86, indicating significant fluctuations around the mean. Future Data Projections indicate that expected values for the time series range between a minimum of 310 and a maximum of 622, with a mean of 0.49 and a standard deviation of 92.75, suggesting increased variability compared to the historical period. The median value is projected to be around 453.34. Anticipated peaks are likely to occur at steps 5, 15, 25, and so on, while predicted dips are expected at steps 8, 18, 28, and similar intervals. Statistical Summary for the future time series reveals the following quartiles: the first quartile (Q1) is **, the median is **, and the third quartile (Q3) is **. Periodicity and Pattern Alignment show consistent high values expected every 10 steps, with auto-correlation anticipated around these intervals, reflecting the cyclical nature observed in the historical data.

\textbf{Iteration 1:} This dataset provides monthly totals of US airline passengers from 1949 to 1960. The prediction length is 29 time steps. **Future Data Description:** Future data projections indicate that expected values for the time series range between a **minimum of 310** and a **maximum of 622**, with a **mean of approximately 440.31** and a **standard deviation of about 79.52**, suggesting variability similar to the historical period. The **median value** is projected to be around **419**. The **first quartile (Q1)** is approximately **390.5**, and the **third quartile (Q3)** is approximately **488.5**. Anticipated **peaks** are likely to occur at **steps 5, 15, and 25**, while predicted **dips** are expected at **steps 8, 18, and 28**, reflecting the cyclical patterns observed in the historical data. **Periodicity and pattern alignment** indicate consistent high values expected every **10 steps**, with **auto-correlation** anticipated around these intervals, continuing the cyclical nature observed previously. 

\textbf{Iteration 2:} Future Data Projections indicate that expected values for the time series range between a minimum of 0.004 and a maximum of 0.98, with a mean of 0.49 and a standard deviation of 0.29, suggesting increased variability compared to the historical period. The median value is projected to be around 0.45. Anticipated peaks are likely to occur at steps 5, 15, 25, and so on, while predicted dips are expected at steps 8, 18, 28, and similar intervals. Statistical Summary for the future time series reveals the following quartiles: the first quartile (Q1) is 0.21, the median is 0.45, and the third quartile (Q3) is 0.70. Periodicity and Pattern Alignment show consistent high values expected every 10 steps, with auto-correlation anticipated around these intervals, reflecting the cyclical nature observed in the historical data.

\textbf{Iteration 3:} The air passengers dataset provides monthly totals of US airline passengers from 1949 to 1960.  Data Projections indicate that expected values for the time series range between a minimum of 310 and a maximum of 622, with a mean of 453.34 and a standard deviation of 92.75, suggesting increased variability compared to the historical period.  Anticipated peaks are likely to occur at steps 5, 15, 25, and so on, while predicted dips are expected at steps 8, 18, 28, and similar intervals.

\textbf{Refined Text:} The air passengers dataset provides monthly totals of US airline passengers from 1949 to 1960. The prediction length is 29 time steps. Data Projections indicate that expected values for the time series range between a minimum of 310 and a maximum of 622, with a mean of 453.34 and a standard deviation of 92.75, suggesting increased variability compared to the historical period.  Anticipated peaks are likely to occur at steps 5, 15, 25, and so on, while predicted dips are expected at steps 8, 18, 28, and similar intervals.

 \subsection{Pipeline for Collect the text candidate}
\label{text_single-agent}

\begin{figure}[htbp]
   \centering
   \includegraphics[width=\linewidth]{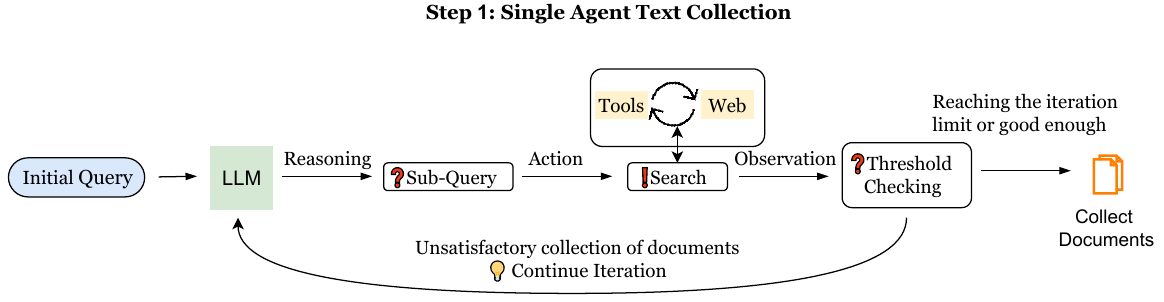}
   \caption{The pipeline of collect the document for extract template. Leverage the ReAct to inspire agent collect human-craft text about time series description. The framework stopped when collect enough document or reach the max iteration limitation.}
   \label{fig:single_agent}
\end{figure}

Figure \ref{fig:single_agent} shows how the single agent framework is proposed how to collect templates. While direct search for relevant content is feasible, it is constrained by a maximum of $K$ titles relevant to the query keyword. To overcome this, we aim to gather relevant candidates based on content similarity. For instance, a simple search for \textit{``time series generation''} might return its definition, but a reasoning-enabled agent can plan what types of articles are more likely to contain relevant content, thereby diversifying the search results. 

\subsection{Detail of Automatic Evaluation}
\label{automatic_evaluation}
The system takes as input a historical TS alongside its textual description, using these as conditions to predict future TS values. This design leverages the intuition that historical TS serves as supplementary contextual information, simplifying the generation process while constraining the model’s output space. This framework isolates the text’s impact on forecasting accuracy, minimizing the influence of the TS data itself to better assess the quality of the textual input. To validate this approach, we conduct evaluations on two widely studied datasets: the \textit{Air Passenger} and \textit{Sunspots} datasets, which offer diverse temporal patterns and domain characteristics. The key to our refined framework lies in the definition of the evaluation dimension, which directly influences the agent’s ability to correct text-time series pair errors and provide high-quality feedback. On one hand, we define evaluation criteria that align with the modal characteristics of both text and TS, allowing the agent to consider both simultaneously in order to correct the data. On the other hand, we also allow the agent to propose more suitable evaluation metrics.

\subsection{Evaluation Dimensions}
\label{dataset_evaluation}
In this section, we detail the initial evaluation criteria and definitions used to assess the generated text and its impact on TS forecasting. These criteria are designed to align with the modal characteristics of both text and time series, enabling the agent to evaluate and correct the input-output pairs effectively. We consider both text and TS metrics. Specifically, we consider following dimensions for text:

\begin{itemize}
    \item \textbf{Accuracy of trend description:} The description accurately identifies the steady increase in the time series.
    \item \textbf{Mention of seasonality:} The description correctly notes the absence of seasonality in the data.
    \item \textbf{Completeness of information:} The description covers the main aspects of the time series but could mention the exact rate of increase.
    \item \textbf{Clarity of description:} The description is clear and easy to understand.
\end{itemize}

we consider following for time series:
Specifically, we consider Mean Squared Error (MSE) \citep{hurvich1988mean}, Kolmogorov-Smirnov Test (K-S Test) \citep{berger2014kolmogorov} and Wasserstein Distance (WD) \citep{panaretos2019statistical} for measuring the difference between the generated and target time series, and building a 5-point Likert scale for evaluate the text quality with 5 dimension (i.e. Accuracy of trend description; Mention of seasonality; Reference to external factors; Clarity of description; Completeness of information).

\subsection{Initial Template Exact from Collect Template}
\label{template_bank}
The time series templates extracted from the collected corpus typically contain descriptions of key patterns such as trends, seasonalities, and changes over time. For example, a typical template could be structured as:

\vspace{-0.5em}
{\small 
\begin{spacing}{0.85}
\textit{~~~~Overall, \{\textcolor{gray}{entity}\}, \{describe\_general\_trend\}. At the beginning, \{\textcolor{gray}{detail\_initial}\}. As time progressed, \{\textcolor{gray}{change\_description}\}, culminating in \{\textcolor{gray}{end\_description}\} by  \{\textcolor{gray}{end\_time}\}}
\end{spacing}
}
\vspace{-0.5em}

Additionally, the templates may include other relevant information, such as statistical metrics (e.g., minimum, maximum, standard deviation), dataset information, degree words (e.g., dramatically, slightly) that describe the intensity of changes, and the time series length."

\subsection{Example of Refined Text Template}
\label{template_list}


{\small
\begin{spacing}{0.85}
\textit{\textnormal{1.}The {\textcolor{gray}{dataset\_name}} dataset provides {\textcolor{gray}{frequency}} totals of {\textcolor{gray}{data\_description}} from {\textcolor{gray}{start\_date}} to {\textcolor{gray}{end\_date}}. The prediction length is {\textcolor{gray}{prediction\_length}} time steps. Data statistics indicate that expected values for the time series range between a minimum of {\textcolor{gray}{min\_value}} and a maximum of {\textcolor{gray}{max\_value}}, with a mean of {\textcolor{gray}{mean\_value}} and a standard deviation of {\textcolor{gray}{std\_value}}, suggesting {\textcolor{gray}{variability\_summary}} compared to the beginning period. Anticipated peaks are likely to occur at steps {\textcolor{gray}{peak\_steps}}, while predicted dips are expected at steps {\textcolor{gray}{dip\_steps}}.}
\end{spacing}
}

{\small
\begin{spacing}{0.85}
\textit{\textnormal{2.}The {\textcolor{gray}{dataset\_name}} dataset captures {\textcolor{gray}{frequency}} measurements of {\textcolor{gray}{data\_description}} collected between {\textcolor{gray}{start\_date}} and {\textcolor{gray}{end\_date}}. Each series spans {\textcolor{gray}{prediction\_length}} steps. Statistical indicators show values ranging from {\textcolor{gray}{min\_value}} to {\textcolor{gray}{max\_value}}, with an average of {\textcolor{gray}{mean\_value}} and a variability (standard deviation) of {\textcolor{gray}{std\_value}}. Notable shifts in the data occur around steps {\textcolor{gray}{peak\_steps}} (highs) and {\textcolor{gray}{dip\_steps}} (lows), illustrating {\textcolor{gray}{variability\_summary}} trends.}
\end{spacing}
}

{\small
\begin{spacing}{0.85}
\textit{\textnormal{3.}With a focus on {\textcolor{gray}{data\_description}}, the {\textcolor{gray}{dataset\_name}} dataset provides {\textcolor{gray}{frequency}} records from {\textcolor{gray}{start\_date}} to {\textcolor{gray}{end\_date}}. Analysis shows values peaking at {\textcolor{gray}{max\_value}} and dipping to a minimum of {\textcolor{gray}{min\_value}}, while maintaining an average of {\textcolor{gray}{mean\_value}}. The standard deviation of {\textcolor{gray}{std\_value}} highlights {\textcolor{gray}{variability\_summary}}. Prediction spans of {\textcolor{gray}{prediction\_length}} time steps reveal anticipated peaks at {\textcolor{gray}{peak\_steps}} and dips at {\textcolor{gray}{dip\_steps}}.}
\end{spacing}
}

{\small
\begin{spacing}{0.85}
\textit{\textnormal{4.}Designed for generation/forecasting, the {\textcolor{gray}{dataset\_name}} dataset offers {\textcolor{gray}{frequency}} intervals of {\textcolor{gray}{data\_description}} over a timeline from {\textcolor{gray}{start\_date}} to {\textcolor{gray}{end\_date}}. Prediction windows of {\textcolor{gray}{prediction\_length}} steps enable users to observe patterns such as peak values at {\textcolor{gray}{peak\_steps}} and dips around {\textcolor{gray}{dip\_steps}}. Ranges from {\textcolor{gray}{min\_value}} to {\textcolor{gray}{max\_value}} suggest considerable variability, with a mean of {\textcolor{gray}{mean\_value}} and standard deviation of {\textcolor{gray}{std\_value}}, illustrating {\textcolor{gray}{variability\_summary}} across the dataset.}
\end{spacing}
}

{\small
\begin{spacing}{0.85}
\textit{\textnormal{5.}The {\textcolor{gray}{dataset\_name}} dataset, spanning {\textcolor{gray}{start\_date}} to {\textcolor{gray}{end\_date}}, provides {\textcolor{gray}{frequency}} observations of {\textcolor{gray}{data\_description}}. Its prediction horizon is set at {\textcolor{gray}{prediction\_length}} steps. Statistical analysis reveals that the data ranges from a low of {\textcolor{gray}{min\_value}} to a high of {\textcolor{gray}{max\_value}}, with a mean of {\textcolor{gray}{mean\_value}} and a variability of {\textcolor{gray}{std\_value}}. Peaks are observed at {\textcolor{gray}{peak\_steps}}, while dips are noticeable around {\textcolor{gray}{dip\_steps}}, demonstrating {\textcolor{gray}{variability\_summary}} throughout the series.}
\end{spacing}
}

{\small
\begin{spacing}{0.85}
\textit{\textnormal{6.}Spanning from {\textcolor{gray}{start\_date}} to {\textcolor{gray}{end\_date}}, the {\textcolor{gray}{dataset\_name}} dataset includes {\textcolor{gray}{frequency}} records of {\textcolor{gray}{data\_description}}. Predictions are made for horizons of {\textcolor{gray}{prediction\_length}} time steps. The series ranges between {\textcolor{gray}{min\_value}} and {\textcolor{gray}{max\_value}}, averaging {\textcolor{gray}{mean\_value}} with variability marked by a standard deviation of {\textcolor{gray}{std\_value}}. Anticipated patterns show higher values near {\textcolor{gray}{peak\_steps}} and lower ones around {\textcolor{gray}{dip\_steps}}, reflecting {\textcolor{gray}{variability\_summary}} over time.}
\end{spacing}
}

{\small
\begin{spacing}{0.85}
\textit{\textnormal{7.}The {\textcolor{gray}{dataset\_name}} dataset includes {\textcolor{gray}{domain}} time series with {\textcolor{gray}{frequency}} frequency, spanning from {\textcolor{gray}{start\_date}} to {\textcolor{gray}{end\_date}}. The time series contain {\textcolor{gray}{total\_steps}} time points, with predictions required for {\textcolor{gray}{prediction\_length}} steps ahead. The dataset's range of values varies between {\textcolor{gray}{min\_value}} and {\textcolor{gray}{max\_value}}, with a mean of {\textcolor{gray}{mean\_value}} and a standard deviation of {\textcolor{gray}{std\_value}}. The series shows notable fluctuations, with potential peaks observed at {\textcolor{gray}{peak\_steps}} and troughs around {\textcolor{gray}{dip\_steps}}.}
\end{spacing}
}

{\small
\begin{spacing}{0.85}
\textit{\textnormal{8.}The {\textcolor{gray}{dataset\_name}} dataset's {\textcolor{gray}{subdataset}} includes {\textcolor{gray}{domain}} time series data collected at {\textcolor{gray}{frequency}} intervals, spanning from {\textcolor{gray}{start\_date}} to {\textcolor{gray}{end\_date}}. The total number of time steps is {\textcolor{gray}{total\_steps}}, and the forecast horizon is {\textcolor{gray}{prediction\_length}}. Data statistics reveal that the values range from a minimum of {\textcolor{gray}{min\_value}} to a maximum of {\textcolor{gray}{max\_value}}, with an average of {\textcolor{gray}{mean\_value}} and a standard deviation of {\textcolor{gray}{std\_value}}, indicating {\textcolor{gray}{variability\_summary}}. Notable peaks are anticipated at time steps {\textcolor{gray}{peak\_steps}}, while predicted dips are expected at {\textcolor{gray}{dip\_steps}}.}
\end{spacing}
}

{\small
\begin{spacing}{0.85}
\textit{\textnormal{9.}The {\textcolor{gray}{subdataset}} from the {\textcolor{gray}{dataset\_name}} dataset contains {\textcolor{gray}{domain}} time series recorded at {\textcolor{gray}{frequency}} intervals, ranging from {\textcolor{gray}{start\_date}} to {\textcolor{gray}{end\_date}}. The series consist of {\textcolor{gray}{total\_steps}} time points, and the forecast length is {\textcolor{gray}{prediction\_length}}. The time series data spans a range of values from {\textcolor{gray}{min\_value}} to {\textcolor{gray}{max\_value}}, with an average of {\textcolor{gray}{mean\_value}} and a standard deviation of {\textcolor{gray}{std\_value}}. The data exhibits {\textcolor{gray}{variability\_summary}}, with potential peaks at {\textcolor{gray}{peak\_steps}} and troughs at {\textcolor{gray}{dip\_steps}}.}
\end{spacing}
}

\subsection{Using LLM directly for Time Series Forecasting}
\label{LLM_directly}

We first explored whether rule-based text could work here as it sample and easy to implement. But its results show fluctuations within a certain range, which confuses the model.

Example rule-based template like \textit{"The air passengers dataset provides monthly totals of US airline passengers from 1949 to 1960. The data can be described as small consolidation small consolidation small consolidation small consolidation small breakout rebound small consolidation small rounded\_bottom small consolidation small downtrend small consolidation small false\_breakout rebound small breakout small downtrend slight breakout small downtrend rebound small breakout moderate downtrend small breakout moderate downtrend small false\_breakout small breakout moderate downtrend pullback small breakout small false\_breakout small double\_bottom small breakout significant downtrend."}

We further leverage LLM to optimize it, aiming to make it more narrative-like. For example

\textit{The input time series illustrates a market with frequent minor fluctuations and distinct pattern formations. It begins with a small consolidation, followed by a false breakout and a rounded bottom. After another phase of consolidation, the market experiences a slight rebound, then consolidates again before another rebound. This is followed by a small breakout and the formation of a double bottom pattern. A rounded bottom and descending triangle emerge, succeeded by a pullback. The market then forms another double bottom and rounded bottom, followed by a period of consolidation. Two more rounded bottoms appear before a small breakout and a subsequent rebound. The series concludes with a final phase of consolidation and the formation of a double bottom.}

\textit{The input time series illustrates a market with frequent minor fluctuations and distinct pattern formations. It begins with a pullback, followed by a small false breakout and a rebound. After a brief period of consolidation, the market experiences another pullback and forms a small rounded bottom. This is followed by another phase of consolidation and a second small rounded bottom, leading to a rebound. The market then forms a small rectangle pattern, rebounds again, and creates a small wedge before pulling back once more. Another small rounded bottom forms, followed by a rebound and two consecutive small false breakouts. The series then undergoes another pullback, experiences a small breakout, forms a small rectangle, and pulls back again. Finally, the market rebounds and concludes with a small breakout.}

Directly employed In-context learning (ICL) to activate LLMs for text generation is also considered. In this setup, the time series first adopts Seasonal-Trend decomposition using Loess (STL) \citep{cleveland1990stl}, which is a robust method to decompose time series into long-term trend, seasonal, and residual components. Then, descriptions are generated separately for the initial, intermediate, final, and overall trends. It is important to note that this textual description is based on periodicity rather than time, as the time series is more nuanced. Descriptions segmented by time showed erroneous outputs in experiments, particularly in the form of regular fluctuations within specific intervals. For detailed prompt design consult. Example like:

**Initial Phase (roughly first 30-35 points)**: \textit{The series starts with relatively low values, mostly fluctuating between 4 and 6.
There's a slight downward trend initially, followed by a period of relative stability.
A few short-lived spikes occur (e.g., around points 7-8), but quickly return to the baseline.}

**Transition Phase (around points 35-40)**: \textit{There's a sudden and significant level shift upwards.
Values jump from the 4-6 range to the 8-10 range.
This marks the most dramatic change in the entire series.}

**Middle Phase (roughly points 40-70)**: \textit{The series settles into a new, higher range, mostly between 9 and 11.
There's increased volatility compared to the initial phase.
A slight upward trend is visible within this phase.
Several cycles of rises and falls occur, but each cycle tends to peak higher than the last.}

**Late Phase (final 30-35 points)**: \textit{The upward trend becomes more pronounced.
Volatility increases further, with larger swings between highs and lows.
The series reaches its highest points in this phase, with peaks above 12.
Despite the higher peaks, there are still significant drops, creating a saw-tooth pattern.}

**Overall Trend**: \textit{The series shows a clear overall upward trend from start to finish.
This trend is not linear but rather step-like, with a major shift in the middle and then a more gradual increase.
The trend is accompanied by increasing volatility over time.}

\begin{figure}
    \centering
    \includegraphics[scale=0.5]{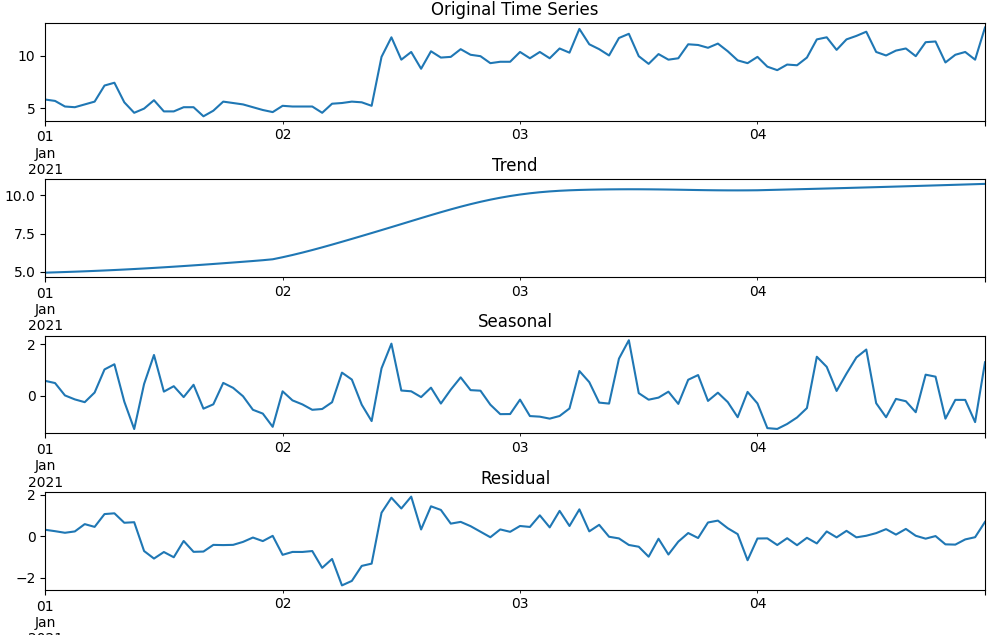}
    \caption{STL Visualization}
    \label{fig:stl}
\end{figure}



\subsection{Evaluation}
To further ensure the neutrality and safety of the text templates, we conducted the verifications by both automatic and human evaluation: We used GPT-4 to assess whether the templates could be linked to specific TS. Results confirmed that no such linkage existed.  Independent human reviewers examined the templates and concluded that they did not reveal any specific details about the TS. 

\section{Diffusion-based Time Series Generation}

\subsection{Semantic Prototype Assignment and Alignment}
\label{Semantic_Prototype_Assignment_Alignment}

\textbf{Semantic Prototype Assignment.} Although the same set of prototypes is used across different instances, the degree to which each prototype explains different instances varies. 
To address this, we assign prototypes to each time series and text description pair, which serves as a condition for the generation model. 
For each input sequence $x$ (comprising both the time series and text embeddings), a weight vector is generated, the dimension of which corresponds to the number of prototypes. This is achieved via a feature extractor $\phi$. 
Each element of the vector $\phi(x)_i$ reflects the contribution of each prototype unit $p_i$ in the prototype set $P$, and these weights modify the attention mechanism used during generation. As a result, the model is conditioned on the assigned weighted prototypes. The weights are applied through an attention mask $m$, which operates on the attention weights for prototypes. To ensure sparsity, we discard prototype units that are assigned with negative weights by setting their attention weights to zero. Formally, the prototype assignments are transformed into attention mask $m$ as follows:
\begin{equation}
     \quad m = \phi(x_0,t_0) - I_{\phi(x_0,t_0)\leq0} \cdot \infty 
\end{equation}

where $\phi(x,t) \in \mathbb{R}^{N \times d}$ is the output from the feature extraction layer that processes both time series and text embeddings. $I_{\phi(x,t)\leq0}$ is an indicator function that zeroes out negative weights, ensuring that only retains non-negative values. In this work, we use randomly orthogonal vectors as prototypes \citep{huang2025timedp}. In future work, we will consider methods such as fine-tuning.

\textbf{Semantic Prototype Alignment.} To condition the denoising diffusion process, we adapt the denoising objective using $c$ as a condition, influencing the model’s intermediate layers through cross-attention. This ensures that the generated time series aligns with the specified conditional instruction. To achieve this, we aim to align the condition and semantic prototypes during the training phase. A set number of query embeddings are allocated to both the text and time series as input. These queries interact with each semantic prototype through cross-attention layers (inserted every other transformer block $z$). We initialise the weights of the cross-attention layers randomly and update them during training. Specifically, we apply cross-attention to the feature representations using the following equations:
\begin{equation}
Q = W_Q \cdot cz, \quad K = W_K \cdot \mathbf{P}, \quad V = W_V \cdot \mathbf{P}
\end{equation}
where
\begin{equation}
z = \text{FF} \left( \text{softmax} \left( \frac{Q (K)^T}{\sqrt{d}} + m \right) \cdot V \right)
\end{equation}

Here, $z \in \mathbb{R}^{N \times d}$ denotes the output from the attention block. $W_Q, W_K, W_V \in \mathbb{R}^{d \times d}$ are learnable projection matrices applied on the sequence dimension. The attention output $z_{\text{final}}$ is passed to another feedforward network to produce the final output $\hat{\epsilon} = \text{FF}(z_{\text{final}})$. 

\subsection{Algorithm for Generation}
\label{algorithm}
Algorithm \ref{alg:algorithm-text_training} and Algorithm \ref{alg:algorithm-text_sampling} shown the diffusion process of generate new sample with text guide.

    
    

\begin{algorithm}[t!]
\caption{Bridge Training}
\label{alg:algorithm-text_training}
\begin{algorithmic}[1] 
    \REQUIRE{Time series-text dataset $\{\mathcal{X,T}\}$}
    \ENSURE{Network parameters $\phi$ and $\theta$, prototypes $P$}
    
    \STATE Initialize prototypes $P$
    
    \REPEAT
        \STATE Sample $x_0, s$ from $\{\mathcal{X, T}\}$
        \STATE Extract prototype assignments $m=\phi(x_0, P)$
        \STATE Randomly set $P$ as unconditional identifier $p_u$
        \STATE Randomly sample time step $n \sim \mathcal{U}(1, N)$
        \STATE Randomly sample noise $\epsilon \sim \mathcal{N}(0, I)$
        \STATE Encode text description: $l = \text{TextEncoder}(s)$
        \STATE Corrupt data $x_n = \sqrt{\alpha_n}x_0 + \sqrt{1-\alpha_n}\epsilon$
        \STATE Predict step noise with $\tilde{\epsilon} = \tilde{\epsilon}_{\theta,P}(x_n, n, m, l)$
        \STATE Compute loss and take gradient step
    \UNTIL{maximum training step}
    
    \STATE \textbf{Output:} Network parameters $\phi$, $\theta$ and prototypes $P$
\end{algorithmic}
\end{algorithm}

\begin{algorithm}[t!]
\caption{Bridge Inference}
\label{alg:algorithm-text_sampling}
\begin{algorithmic}[1] 
    \REQUIRE{Prototypes $P$, time series prompts $x$ from train set or few-shot demonstration set, and text description $s$ from test set}
    \ENSURE{Generated time series samples $\hat{x}$}
    
    \STATE Extract prototype prompts $m=\phi(x, P)$
    \STATE Encode text description $l = \text{TextEncoder}(s)$
    \STATE Randomly sample noise $\hat{x}_N \sim \mathcal{N}(0, I)$
    
    \FOR{$n$ from $N$ to $1$}
        \STATE Predict step noise with $\tilde{\epsilon}_n = \tilde{\epsilon}_{\theta,P}(\hat{x}_n, n, m, l)$
        \STATE Denoise $\hat{x}_{n-1} = \frac{\hat{x}_n - \sqrt{1-\bar{\alpha}_n}\tilde{\epsilon}_n}{\sqrt{\alpha_n}}$
    \ENDFOR
    
    \STATE $\hat{x} = \hat{x}_0$
    \STATE \textbf{Output:} Generated samples $\hat{x}$
\end{algorithmic}
\end{algorithm}

\subsection{The form of Input and Output}
\label{formofinputandoutput}
Input of Diffusion Model including a time series and corresponding text description.  The two of them will be processed by the encoder and LLama respectively, and the obtained embedding will be fused through a single-layer MLP as conditional input. The output of the diffusion model is a synthetic time series.

\section{Baseline Model}
\subsection{Time Series Generation Model}
\label{TS_generation_model}

\textbf{TimeVQVAE}\citep{DBLP:conf/aistats/LeeMA23} is a generative model designed for sequential data. It combines the strengths of a variational autoencoder (VAE) with vector quantisation to discretise latent space representations, making it effective for time series data. The model consists of an encoder that compresses the input data into a discrete latent space and a decoder that reconstructs the time series. TimeVQVAE is particularly useful for generating realistic time series samples while maintaining key temporal dependencies. The quantisation step helps in learning discrete representations that can be reused for efficient time series modelling and generation.

\textbf{TimeGAN} \citep{DBLP:conf/nips/YoonJS19} is a variant of the GAN framework specifically tailored for time series data. It combines both supervised and unsupervised learning approaches, using a generator to create synthetic time series and a discriminator to differentiate between real and generated data. Additionally, it integrates an embedding network to capture temporal dependencies and preserve temporal correlations between generated samples. The model ensures that the generated time series not only closely mimic the statistical properties of the original data but also maintain the correct temporal ordering and dynamics. TimegGAN is particularly useful in applications requiring realistic synthetic data generation, such as forecasting and anomaly detection.

\textbf{GT-GAN} \citep{DBLP:conf/nips/JeonKSCP22} introduces a novel architecture for time series generation by incorporating both global and local perspectives. The model features two generators: one focuses on capturing the global trends across the entire time series, while the other focuses on local variations. The two components work together to ensure that the generated time series exhibit realistic patterns on both macro and micro levels. GT-GAN uses a two-stream discriminator that evaluates both the global and local outputs, ensuring high fidelity in the generated data. This model is effective for generating complex time series where both long-term trends and short-term fluctuations are important.

\textbf{TimeVAE} \citep{DBLP:journals/corr/abs-2111-08095} extends the traditional VAE architecture to model time series data. It uses an encoder to map time series data into a continuous latent space, from which the decoder reconstructs the original time series. The model captures uncertainty and variation in the data through the latent space’s probabilistic structure, making it well-suited for applications where capturing latent factors and generating multiple plausible future scenarios is important. TimeVAE can be applied to various tasks, such as anomaly detection, forecasting, and data augmentation, by learning complex temporal dependencies and generating realistic time series that adhere to the original data’s statistical properties.

\subsection{Time Series Forecasting Model}
\label{TS_forecasting_model}

\textbf{Time-LLM} \citep{DBLP:journals/corr/abs-2310-01728} is a powerful TS LLM that outperforms specialized forecasting models, which repurposes LLMs for time series forecasting by reprogramming input data and employing the Prompt-as-Prefix (PaP) technique for enhanced context alignment.

\textbf{GPT4TS} \citep{DBLP:conf/nips/ZhouNW0023} takes advantage of pre-trained language and vision models for general time series analysis. By demonstrating that supervised fine-tuning (SFT) can successfully extend LLM capabilities to time series tasks, GPT4TS bridges the gap between natural language processing models and temporal data analysis. The model’s architecture shows the feasibility of applying large pre-trained models to time series, leading to significant performance improvements in various time series applications.

\textbf{LLM4TS} \citep{DBLP:journals/corr/abs-2308-08469} is an innovative framework that repurposes pre-trained LLMs for time-series forecasting, employing a two-stage fine-tuning strategy and a two-level aggregation method to align with and enhance the model's ability to process multi-scale temporal data, outperforming state-of-the-art models in both fune-tuing and few-shot scenarios. 

\textbf{TEMPO} \citep{DBLP:conf/iclr/CaoJAPZY024} proposed using prompts to adapt to different time series distributions. It demonstrates superior performance in zero-shot settings across diverse benchmark datasets, showcasing its potential as a foundational model-building framework for capturing dynamic temporal phenomena.

\section{Experiment Setup}
\label{experiment_setup}
The time series length \( T \) for generation is set to 168 in a form of non-overlap sequence slices for all the datasets. For forecasting, we assessed performance over four different prediction horizons \( H \in \{24, 36, 48, 60\} \) for ILI and \( H \in \{6, 48\} \) for M4.   The dataset is divided into training, validation, and test sets in a ratio of 8:1:1. All models are tested with the same dataset size, as adding text descriptions to time series sliding is resource-intensive, prompting a reduction in the number of sliding windows.

\section{Implementation Detail}
\label{implementation_detail}
We implemented all the model and conduct all experiments on  single NVIDIA Tesla A100 80GB GPUs. For LLM used in proposed model is LLama3-8B \citep{DBLP:journals/corr/abs-2407-21783}. For generation task, we keep all model's sequence length is 168 which is the max length of \texttt{Pedestrian}, \texttt{Rain}, \texttt{Temperature} datasets. For evaluation of the synthesis data quality task, we keep the sequence length of 256. The dataset is divided into training, validation, and test sets in a ratio of 8:1:1.

The reported result are all under following training settings. The number of prototypes are set to 16 for all the main evaluations. Models for each sequence length are
trained for 50, 000 steps using a batch size of 128 and a
learning rate of $5 * 10^{-5}$ with 1, 000 warm-up steps.

\section{Dataset Analysis}
\label{dataset_statistics}
\subsection{Details of Datasets}
In this section, we provide a detailed overview of the datasets used for model training in this paper:

\textbf{Electricity:} This dataset represents the hourly electricity consumption of 321 clients from 2012 to 2014, measured in kilowatts (kW). It was originally extracted from the UCI repository.

\textbf{Solar:} This dataset contains 137 time series representing hourly solar power production in the state of Alabama throughout 2006.

\textbf{Wind:} This dataset contains a single extensive daily time series representing wind power production (in megawatts) recorded at 4-second intervals starting from August 1, 2019. It was downloaded from the Australian Energy Market Operator (AEMO) online platform.

\textbf{Traffic:} This dataset contains 15 months of daily data (440 daily records) describing the occupancy rate (between 0 and 1) of different car lanes on San Francisco Bay Area freeways over time.

\textbf{Taxi:} This dataset contains spatio-temporal traffic time series of New York City taxi rides recorded at 1,214 locations every 30 minutes during January 2015 and January 2016.

\textbf{Pedestrian:} This dataset contains hourly pedestrian counts captured from 66 sensors in Melbourne starting from May 2009. The original dataset is regularly updated when new observations become available. The dataset used here contains pedestrian counts up to April 30, 2020.

\textbf{Air Quality:} This dataset was used in the KDD Cup 2018 forecasting competition. It contains hourly air quality measurements from 59 stations in two cities: Beijing (35 stations) and London (24 stations) from January 1, 2017, to March 31, 2018. The air quality measurements include various metrics such as PM2.5, PM10, NO2, CO, O3, and SO2. Missing values were imputed using leading zeros or the Last Observation Carried Forward (LOCF) method.

\textbf{Temperature:} This dataset contains 32,072 daily time series with temperature observations and rain forecasts gathered by the Australian Bureau of Meteorology from 422 weather stations across Australia between May 2, 2015, and April 26, 2017. Missing values were replaced with zeros, and the mean temperature column was extracted for use.

\textbf{Rain:} This dataset focuses on rain data extracted from the same source as the Temperature dataset.

\textbf{NN5:} This dataset was used in the NN5 forecasting competition. It contains 111 time series from the banking domain with the goal of predicting daily cash withdrawals from ATMs in the UK. Missing values were replaced by the median across all corresponding days of the week throughout the entire series.

\textbf{Fred-MD:} This dataset contains 107 monthly time series reflecting various macroeconomic indicators sourced from the Federal Reserve Bank's FRED-MD database. The series have been differenced and log-transformed following established practices in the literature.

\textbf{Exchange:} This dataset records daily exchange rates for eight currencies.

\textbf{Stock:} This dataset consists of daily stock prices for the symbol GOOG, which is listed on NASDAQ.

\textbf{Web:} This dataset was used in the Kaggle Wikipedia Web Traffic forecasting competition. It contains 145,063 daily time series representing the number of hits or web traffic for Wikipedia pages from July 1, 2015, to September 10, 2017. Missing values were replaced with zeros.

\subsection{Dataset Statistics}
\label{statistics_dataset}

To test the quality of the synthetic data generated by our proposed model, we conducted tests on two additional datasets. In the experiments, we trained the synthetic data to be the same as the original data and tested it on the real datasets. The statistics of the datasets are in Table \ref{tab:dataset_comparison}:

\begin{table}
\centering
\scalebox{0.72}{
\begin{tabular}{ccccccc}
\toprule
Domain & Tasks & Datasets & Dim. & Series Length & Dataset Size & Frequency \\
\midrule
Long-Term& ILI & 7 & {24, 36, 48, 60} & (617, 74, 170) & 1 week & Illness \\
\midrule
\multirow{6}{*}{\begin{tabular}[c]{@{}c@{}}Short-term\\Forecasting\end{tabular}} 
& M4-Yearly & 1 & 6 & (23000, 0, 23000) & Yearly & Demographic \\
& M4-Quarterly & 1 & 8 & (24000, 0, 24000) & Quarterly & Finance \\
& M4-Monthly & 1 & 18 & (48000, 0, 48000) & Monthly & Industry \\
& M4-Weekly & 1 & 13 & (359, 0, 359) & Weekly & Macro \\
& M4-Daily & 1 & 14 & (4227, 0, 4227) & Daily & Micro \\
& M4-Hourly & 1 & 48 & (414, 0, 414) & Hourly & Other \\
\bottomrule
\end{tabular}
}
\caption{Comparison of datasets for long-term and short-term forecasting tasks}
\label{tab:dataset_comparison}
\end{table}


\section{Evaluation Metrics}
\label{baseline_evaluation}

The calculations of these metrics are as follows:

\begin{align*}
\text{MSE} &= \frac{1}{H} \sum_{h=1}^{H} (Y_h - \hat{Y}_h)^2, &
\text{MAE} &= \frac{1}{H} \sum_{h=1}^{H} |Y_h - \hat{Y}_h|, \\
\text{SMAPE} &= \frac{200}{H} \sum_{h=1}^{H} \frac{|Y_h - \hat{Y}_h|}{|Y_h| + |\hat{Y}_h|}, &
\text{MAPE} &= \frac{100}{H} \sum_{h=1}^{H} \frac{|Y_h - \hat{Y}_h|}{|Y_h|}, \\
\text{MASE} &= \frac{1}{H} \sum_{h=1}^{H} \frac{|Y_h - \hat{Y}_h|}{\frac{1}{H-s} \sum_{j=s+1}^{H} |Y_j - Y_{j-s}|}, &
\text{OWA} &= \frac{1}{2} \left( \frac{\text{SMAPE}}{\text{SMAPE}_{\text{Naïve2}}} + \frac{\text{MASE}}{\text{MASE}_{\text{Naïve2}}} \right),
\end{align*}

where $s$ is the periodicity of the time series data, $H$ denotes the number of data points (i.e., prediction horizon in our cases), and $Y_h$ and $\hat{Y}_h$ are the $h$-th ground truth and prediction, where $h \in \{1, \dots, H\}$.

For generation, we consider Marginal Distribution Difference (MDD):
\[
\text{MDD}(P, Q) = \sum_{x \in X} | P(x) - Q(x) |
\]

where $P$ and $Q$ represent the marginal distributions of the real and synthetic data, and $X$ denotes the set of possible values for the variable being analyzed.

Also Kullback-Leibler divergence (K-L) 

\[
D_{KL}(P \| Q) = \sum_{x \in X} P(x) \log\left(\frac{P(x)}{Q(x)}\right)
\]

where $P$ and $Q$ are the two probability distributions being compared, and $X$ represents the set of possible values.

\section{Human Evaluation}
\label{human_evaluation}

In addition to the quantitative metrics, we conducted a human evaluation to assess the quality of the generated time series descriptions. 

\subsection{Evaluation Process:}
The human evaluation involved a team of trained annotators who were tasked with reviewing the generated time series in relation to the given text descriptions and ranking the candidate. The process was organized into two primary stages:

\textbf{Preliminary Setup}
Annotators were presented with a single text description accompanied by a set of time series. Each set was evaluated independently, with annotators assessing how well each time series captured the trends, patterns, and anomalies described in the corresponding text. We define two evaluation settings: \emph(i) HE-Rank, annotators ranked the time series generated by different model settings together with the ground truth, focusing purely on relevance. \emph(ii) HE-Mixed, the same sequences were evaluated alongside three distractor time series randomly sampled from the test set, forming a mixed pool of candidates for relative ranking.

\textbf{Evaluation Criteria}
To help annotators better understand and standardize the evaluation process, we provide a set of reference dimensions.

\begin{itemize}
    \item Relevance to given text descriptions: Whether the time series description accurately reflects the patterns, trends, and anomalies observed in the text.
    \item Semantic Alignment: Whether the time series properly conveys the underlying meaning of the texr, such as identifying upward and downward trends, spikes, and seasonal behaviors.
    \item Plausibility: Whether the time series makes sense within the domain context (e.g., finance, healthcare), and offers a reasonable interpretation of the data.
    \item Coherence: Whether the time series is logically consistent within itself, avoiding contradictions and ensuring that the time series aligns with the observed trends or patterns throughout the text.
\end{itemize}

\textbf{Reliability of Evaluation}
To ensure consistency and minimize evaluator bias, multiple annotators assessed each time series. The final score for each description was calculated as the average score from all annotators. Annotators were trained on a common set of guidelines to ensure that the evaluation criteria were applied consistently across all descriptions.

\section{Performance of downstream tasks}
\label{downstream}

\begin{table*}[t]
    \centering    
    \scalebox{0.65}{
    \begin{tabular}{llccccccccccccccc}
        \toprule
        & \multirow{2}{*}{Dataset} & \multicolumn{3}{c}{Random} & \multicolumn{3}{c}{LLM4TS} & \multicolumn{3}{c}{TEMPO} & \multicolumn{3}{c}{Time-LLM} & \multicolumn{3}{c}{GPT4TS}\\
        & {} & \multicolumn{3}{c}{\makebox[0pt]{MSE \hspace{1cm} MAE}} & \multicolumn{3}{c}{\makebox[0pt]{MSE \hspace{1cm} MAE}} & \multicolumn{3}{c}{\makebox[0pt]{MSE \hspace{1cm} MAE}} & \multicolumn{3}{c}{\makebox[0pt]{MSE \hspace{1cm} MAE}} & \multicolumn{3}{c}{\makebox[0pt]{MSE \hspace{1cm} MAE}}\\
        \cmidrule(lr){3-5} \cmidrule(lr){6-8} \cmidrule(lr){9-11} \cmidrule(lr){12-14} \cmidrule(lr){15-17}
        \multirow{3}{*}{ILI} & \textit{Bridge} & \multicolumn{3}{c}{\multirow{3}{*}{\makebox[0pt]{$8.12$ \hspace{1cm} $2.14$}}}  & \multicolumn{3}{c}{\makebox[0pt]{$1.98$ \hspace{1cm} $0.89$}} & \multicolumn{3}{c}{\makebox[0pt]{$1.21$ \hspace{1cm} $1.02$}} & \multicolumn{3}{c}{\makebox[0pt]{$2.20$ \hspace{1cm} $1.44$}} & \multicolumn{3}{c}{\makebox[0pt]{$2.19$ \hspace{1cm} $1.02$}}\\
        & \textit{Real} & \multicolumn{3}{c}{} & \multicolumn{3}{c}{\makebox[0pt]{$1.86$ \hspace{1cm} $0.86$}}  & \multicolumn{3}{c}{\makebox[0pt]{$0.96$ \hspace{1cm} $0.82$}} & \multicolumn{3}{c}{\makebox[0pt]{$2.00$ \hspace{1cm} $1.20$}} & \multicolumn{3}{c}{\makebox[0pt]{$1.90$ \hspace{1cm} $0.90$}} \\
        & \textit{KernelSynth} & \multicolumn{3}{c}{} & \multicolumn{3}{c}{\makebox[0pt]{$4.35$ \hspace{1cm} $1.50$}}  & \multicolumn{3}{c}{\makebox[0pt]{$1.64$ \hspace{1cm} $1.07$}} & \multicolumn{3}{c}{\makebox[0pt]{$1.43$ \hspace{1cm} $1.01$}} & \multicolumn{3}{c}{\makebox[0pt]{$3.80$ \hspace{1cm} $1.42$}} \\
        \midrule
        & & SMAPE & MASE & OWA & SMAPE & MASE & OWA & SMAPE & MASE & OWA & SMAPE & MASE & OWA & SMAPE & MASE & OWA \\ 
        \cmidrule(lr){3-5} \cmidrule(lr){6-8} \cmidrule(lr){9-11} \cmidrule(lr){12-14} \cmidrule(lr){15-17}
        \multirow{3}{*}{M4}& \textit{Bridge}  & \multirow{3}{*}{$26.46$} & \multirow{3}{*}{$4.43$} & \multirow{3}{*}{$1.92$} & 13.10 & 1.99 & 0.93 & 12.25 & 1.73 & 0.87 & 12.85 & 1.90 & 0.96 & 13.10 & 1.99 & 0.93\\
        &\textit{Real} &  &  &  & $12.33$ & $1.80$ & $0.88$ & $12.09$ & $1.72$ & $0.85$ & $12.50$ & $1.77$ & $0.89$ & $12.70$ & $1.94$ & $0.91$ \\
        &\textit{KernelSynth}  &  &  &  & 14.39 & 2.09 & 1.02 & 13.94 & 2.00 & 0.99 & 12.594 & 1.735 & 0.918 & 14.56 & 2.08 & 1.02  \\
        \bottomrule
    \end{tabular}
    } 
    \caption{Comparison of MSE and MAE for TS forecasting across methods, with results for four forecasting horizons: $H$ $\in$ $\{24,36,48, 60\}$ for ILI and $H$ $\in$ $\{6, 48\}$ for M4. Average results are reported. }
    \label{tab:longtermforecasting_summary}
    \vskip -0.1in
\end{table*}
Table~\ref{tab:longtermforecasting_summary} shows the quality of generated data for the purposes of training models for downstream tasks. We generated synthetic data on two additional datasets to assist existing SOTA models in TS forecasting. All models were trained either using only real data or synthetic data and then tested on real test sets. The results  indicate that training with only synthetic data can achieve comparable performance to real data across all models, as performance differences between real and synthetic data are less visible than differences in performance between architectures. This suggests that the generated data is sufficiently realistic, potentially allowing to share synthesised surrogates of otherwise sensitive data. For comparison, we also employed KernelSynth \citep{DBLP:journals/corr/abs-2403-07815} methods. Both methods effectively provided valuable synthetic data (compared to completely random data), but our proposed approach produced data that more closely resembles real data. This underscores its potential for generating meaningful synthetic data across domains.

\section{Ablation Experiment on the Influence of Text Types on Diffusion Models}
\label{texttype_diffusion}

Table~\ref{tab:mdd_kl_text_types} show the performance of different text types. Overall, the inclusion of text descriptions in time series modeling demonstrates consistent benefits across most datasets, with specific nuances depending on the approach and dataset characteristics. Across most datasets, the inclusion of textual descriptions significantly improves alignment metrics, with the Bridge (w/o Background) approach often demonstrating superior performance compared to both Bridge (w/o Pattern+Statistic) and Bridge (w/o Text).

For MDD, Bridge (w/o Background) consistently outperforms or matches other methods in several datasets. For instance, in the ``Rain'' dataset, Bridge (w/o Background) achieves the lowest score ($5.477$), outperforming Bridge (w/o Pattern+Statistic) ($5.499$) and Bridge (w/o Text) ($6.002$). Similarly, in the ``Pedestrian'' dataset, Bridge (w/o Background) achieves the best score ($0.560$), indicating its ability to capture intricate patterns better than the other methods. While Bridge (w/o Pattern+Statistic) occasionally achieves slightly better results (e.g., in the ``Electricity'' dataset with $0.110$ compared to $0.139$ for Bridge (w/o Background)), Bridge (w/o Background) demonstrates more consistent performance across diverse domains.

The K-L Divergence results further highlight the strengths of Bridge (w/o Background). In the ``Wind'' dataset, for example, Bridge (w/o Background) achieves the lowest divergence ($0.048$), outperforming both Bridge (w/o Pattern+Statistic) ($0.052$) and Bridge (w/o Text) ($0.056$). Similar trends are observed in the ``NN5'' and ``Temperature'' datasets, where Bridge (w/o Background) consistently produces the lowest K-L scores, indicating its capability to effectively encode temporal patterns with textual assistance. Notably, in datasets like ``Air'', Bridge (w/o Background) achieves significant improvements compared to Bridge (w/o Text) , further emphasizing its advantage in handling complex and noisy time series data.

\begin{table*}[htbp]
    \centering
    \scalebox{1}{
    \begin{tabular}{llcccc}
    \toprule
      & Dataset  & w/o Pattern+Statistic & w/o Background & w/o Text & Rule-based\\
    \midrule
      \multirow{12}{*}{\rotatebox[origin=c]{90}{\footnotesize{Marginal Distribution Distance}}} & Electricity & $0.110$ & $0.139$ & $0.135$  & $0.256$\\
      & Solar  & $375.531$ & $375.530$ & $375.531$ & $377.232$\\
      & Wind  & $0.344$ & $0.316$ & $0.304$ & $0.422$\\
      & Traffic  & $0.324$ & $0.309$ & $0.315$ & $1.178$\\
      & Taxi  & $0.328$ & $0.325$ & $0.338$ & $0.641$\\
      & Pedestrian & $0.584$ & $0.560$ & $0.576$ & $1.277$\\
      & Air & $0.472$ & $0.440$ & $0.418$ & $0.665$ \\
      & Temperature  & $0.332$ & $0.331$ & $0.356$ & $0.572$\\
      & Rain  & $5.499$ & $5.477$ & $6.002$ & $9.533$\\
      & NN5  & $0.591$ & $0.570$ & $0.613$ & $1.377$\\ 
      & Fred-MD  & $0.239$ & $0.226$ & $0.228$ & $0.460$\\ 
      & Exchange  & $0.315$ & $0.316$ & $0.376$ & $0.430$\\ 
      \midrule
      \multirow{12}{*}{\rotatebox[origin=c]{90}{\footnotesize{K-L Divergence}}} & Electricity & $0.002$ & $0.003$ & $0.001$ & $0.008$\\
      & Solar  & $0.008$ & $0.006$ & $0.005$ & $0.033$\\
      & Wind  & $0.052$ & $0.048$ & $0.056$ & $0.144$\\
      & Traffic  & $0.013$ & $0.012$ & $0.016$ & $0.027$\\
      & Taxi  & $0.020$ & $0.014$ & $0.020$ & $0.093$ \\
      & Pedestrian  & $0.006$ & $0.009$ & $0.009$ & $0.079$\\
      & Air & $0.006$ & $0.005$ & $0.021$ & $0.042$\\
      & Temperature  & $0.023$ & $0.016$ &  $0.020$ & $0.904$\\
      & Rain & $0.006$ & $0.006$ & $0.008$ & $0.016$\\
      & NN5  & $0.008$ & $0.004$ & $0.004$ & $0.101$\\
      & Fred-MD  & $0.005$ & $0.004$ & $0.023$ & $0.108$\\
      & Exchange  & $0.062$ & $0.057$ & $0.067$ & $0.350$ \\
    \bottomrule
    \end{tabular}
    }
    \caption{The performance of different text type. Marginal distribution distance scores (MDD) and K-L divergence (K-L) are reported.}
    \label{tab:mdd_kl_text_types}
\end{table*}

\section{The Impact of Prototype}
\label{prototype}

The results in Table \ref{tab:ablation_prototypes} demonstrate the influence of prototype quantity on marginal distribution distance and K-L divergence across multiple datasets. Increasing the number of prototypes generally leads to improved performance, as observed in the Electricity dataset, where the marginal distribution distance decreases from 0.615 (4 prototypes) to 0.135 (16 prototypes). Similar trends are evident in datasets such as Traffic (1.211 to 0.315) and NN5 (1.550 to 0.613).

For K-L divergence, a higher number of prototypes often results in lower divergence values, indicating better alignment with the target distribution. For example, in the Taxi dataset, K-L divergence drops from 0.154 (4 prototypes) to 0.003 (16 prototypes). However, certain datasets, such as Wind and Exchange, exhibit less consistent trends, suggesting potential variations in data characteristics affecting prototype effectiveness.

\begin{table*}[htbp]
    \centering  
        \scalebox{0.9}{
            \begin{tabular}{llccccclccccc}
                \toprule
                & Prototypes & 4 & 8 & 16 & 32 & 64 & {} & 4 & 8 & 16 & 32 & 64 \\
                \midrule
                \multirow{12}{*}{\rotatebox[origin=c]{90}{\footnotesize{Marginal Distribution Distance}}} & Electricity & $0.615$ & $0.368$ & $0.135$ & $0.117$ & $0.236$ & \multirow{12}{*}{\rotatebox[origin=c]{90}{\footnotesize{K-L Divergence}}}  & $0.006$ & $0.027$ & $0.001$ & $0.001$ & $0.020$ \\
                & Solar  & $375.536$ & $375.531$ & $375.531$ & $375.532$ & $375.556$ &  & $0.025$ & $0.016$ & $0.005$ & $0.008$ & $0.017$ \\
                & Wind  & $0.271$ & $0.299$ & $0.304$  & $0.314$ & $0.309$ &  & $0.059$ & $0.074$ & $0.056$ & $0.075$ & $0.093$ \\
                & Traffic  & $1.211$ & $0.287$ & $0.315$ & $0.349$ & $0.411$ &  & $0.200$ & $0.018$ & $0.021$ & $0.022$ & $0.027$ \\
                & Taxi  & $1.008$ & $0.433$ & $0.338$ & $0.371$ & $0.384$ &  & $0.154$ & $0.013$ & $0.003$ & $0.020$ & $0.019$ \\
                & Pedestrian  & $1.599$ & $0.921$ & $0.576$ & $0.554$ & $0.552$ &  & $0.075$ & $0.022$ & $0.009$ & $0.003$ & $0.002$ \\
                & Air  & $0.611$ & $0.393$ & $0.418$ & $0.515$ & $0.544$ &  & $0.010$ & $0.009$ & $0.005$ & $0.006$ & $0.009$ \\
                & Temperature  & $0.487$ & $0.317$ & $0.356$ & $0.330$ & $0.307$ &  & $0.113$ & $0.025$ & $0.020$ & $0.017$ & $0.042$ \\
                & Rain  & $5.763$ & $4.981$ & $6.002$ & $5.548$ & $6.420$ &  & $0.014$ & $0.009$ & $0.010$ & $0.008$ & $0.018$ \\
                & NN5  & $1.550$ & $0.796$ & $0.613$ & $0.616$ & $0.573$ &  & $0.130$ & $0.118$ & $0.004$ & $0.005$ & $0.009$\\
                & Fred-MD  & $0.407$ & $0.241$ & $0.228$ & $0.245$ & $0.346$ &  & $0.012$ & $0.015$ & $0.006$ & $0.009$ & $0.030$ \\
                & Exchange  & $0.365$ & $0.359$ & $0.376$ & $0.309$  & $0.330$ &  & $0.062$ & $0.063$ & $0.067$ & $0.083$ & $0.046$ \\
                \bottomrule
            \end{tabular}
        }
        \caption{Ablation experiment on the impact of the number of prototypes. }
        \label{tab:ablation_prototypes}
\end{table*}

\section{The Impact of LLms on the diffusion model performance}
\label{llm_impact}

The Table \ref{tab:llama_gpt2_performance} compares the performance of Llama and GPT2 as encoders in our diffusion model across various time series domains. Both models show similar performance in most domains, with slight differences in specific cases. For example, Llama performs slightly better in the ``Electricity'' ($0.139$ vs $0.174$) and ``NN5'' ($0.570$ vs $0.887$) domains, suggesting a better ability to capture fluctuations in these time series. In contrast, GPT2 outperforms Llama in ``Pedestrian'' ($0.578$ vs $0.483$) and ``Fred-MD'' ($0.226$ vs $0.225$), indicating its strength in encoding gradual trends. Overall, both models show strong performance across multiple domains, with only minor variations. These results highlight that while Llama and GPT2 differ slightly in their handling of specific time series patterns, both are effective encoders for our diffusion model, capable of capturing both domain-specific and general temporal features.

\begin{table}[ht]
\centering
\begin{tabular}{lcccccc}
\toprule
\textbf{Model} & \textbf{Electricity} & \textbf{Solar} & \textbf{Wind} & \textbf{Traffic} & \textbf{Taxi} & \textbf{Pedestrian} \\ 
\midrule
Llama & 0.139 & 375.531 & 0.316 & 0.309 & 0.325 & 0.578 \\ 
GPT2 & 0.174  & 375.538 & 0.325 & 0.331 & 0.361 & 0.483 \\ 
\midrule
\textbf{} & \textbf{Air} & \textbf{Temperature} & \textbf{Rain} & \textbf{NN5} & \textbf{Fred-MD} & \textbf{Exchange} \\ 
\midrule
Llama & 0.440 & 0.331 & 5.932 & 0.570 & 0.226 & 0.374 \\ 
GPT2 & 0.645 & 0.349 & 5.994 & 0.887 & 0.225 & 0.414 \\ 
\bottomrule
\end{tabular}
\caption{Model performance across different domains. Result measured by MDD}
\label{tab:llama_gpt2_performance}
\end{table}

\section{Data Augmentation Results}
\label{da_result}

For long-term forecasting (Table \ref{tab:longtermforecasting}), we find that  the LLM4TS trained via the synthetic data produces relatively low MSE and MAE values, such as ILI-24 Synthesis with an MSE of 1.84 and an MAE of 0.85, which are competitive with the performance on real-world datasets.  In fact, for length like 24 and 36, LLM4TS consistently performs well, showing competitive results in both MSE and MAE, even when compared to training on real data. GPT4TS and Time-LLM, on the other hand, exhibit a slight drop in performance when trained on synthetic data, but considerable accepted. In the short-term forecasting scenario (Table \ref{tab:m4_shorttermforecasting}), the results show similar trends. This suggests that synthetic data can effectively simulate real data patterns, making it a viable option for model training when real-world data is limited or unavailable.

We also compare the quality of our synthetic data generation with the Kernel Synth method employed by Chronos~\cite{DBLP:journals/corr/abs-2403-07815}. While on the long-term ILI forecasting task models trained on our data clearly outperform models trained on KernelSynth, the picture is slightly more nuanced for shorty-term M4 forecasting. Since the forecast horizons are shorter in M4, the overall difference is less nuanced. Table~\ref{tab:significance-tests} details for which sub-datasets and models trained on our data are performing better and for which sub-datasets and models KernelSynth is more suitable. These comparisons are further contextualised by statistical analysis - for SMAPE and MASE metrics, we conduct t-test on the individual scores directly. For OWA, since it involves averages of SMAPE and MASE, we use the bootstrapping technique by sampling $1000$ times with replacement to obtain the distributions of scores and conduct the t-test on this subset. Overall our method performs better on half model/subset combinations, of which all but two results are statistically significant. 

\begin{table*}[h]
    \centering
    \scalebox{1}{
    \begin{tabular}{l|cc|cc|cc|cc}
        \toprule
        Methods & \multicolumn{2}{c}{LLM4TS}  & \multicolumn{2}{c}{TEMPO} & \multicolumn{2}{c}{Time-LLM} & \multicolumn{2}{c}{GPT4TS}\\
        \cmidrule(lr){2-3} \cmidrule(lr){4-5} \cmidrule(lr){6-7} \cmidrule(lr){8-9} 
        Metrics & MSE & MAE & MSE & MAE & MSE & MAE & MSE & MAE\\
        \midrule
       \textit{ILI-24 KernelSynth}  & 4.36 & 1.49 & 1.48 & 1.02 & 1.21 & 0.93 & 3.92 & 1.45 \\
       \textit{ILI-36 KernelSynth}  & 4.32 & 1.49 & 1.37 & 0.96 & 1.31 & 0.94 & 3.87 & 1.43  \\
       \textit{ILI-48 KernelSynth}  & 4.15 & 1.48 & 1.69 & 1.09 & 1.49 & 1.04 & 3.77 & 1.40 \\
       \textit{ILI-60 KernelSynth}  & 4.35 & 1.50 & 2.01 & 1.22 & 1.71 & 1.11 & 3.62 & 1.39 \\
       \midrule
        \textit{ILI-24 Ours} & 1.84 & 0.85 & 1.00 & 0.87 & 2.05 & 1.29 & 2.23 & 0.99 \\
        \textit{ILI-36 Ours} & 1.86 & 0.86 & 1.22 & 0.99 & 2.13 & 1.34 & 2.13 & 0.97\\
        \textit{ILI-48 Ours}  & 1.88 & 0.88 & 1.34 & 1.08 & 2.35 & 1.60 & 2.28 & 1.05 \\
        \textit{ILI-60 Ours}  & 2.37 & 0.99 & 1.49 & 1.14 & 2.30 & 1.55 & 2.35 & 1.09 \\
        
        \midrule
        \textit{ILI-24 Real} & 1.78 & 0.81 & 0.66 & 0.63 & 1.83 & 1.15 & 1.99 & 0.88\\
        \textit{ILI-36 Real} & 1.75 & 0.82 & 0.92 & 0.80 & 1.90 & 1.17 & 1.90 & 0.90\\
        \textit{ILI-48 Real} & 1.72 & 0.84 & 1.33 & 1.02 & 2.16 & 1.26 & 1.81 & 0.88\\
        \textit{ILI-60 Real} & 2.20 & 0.95 & 0.91 & 0.80 & 2.11 & 1.23 & 1.87 & 0.92\\
        \bottomrule
    \end{tabular}
    }
    \caption{Comparison of MSE and MAE across various methods on Long-term forecasting. The results are for four different forecasting horizons: $H$ $\in$ $\{24,36,48, 60\}$. Red values indicate the best score, and blue values represent the second best.}
    \label{tab:longtermforecasting}
\end{table*}

\begin{table*}[h]
    \centering
    \scalebox{0.53}{
    \begin{tabular}{l|ccc|ccc|ccc|ccc|ccc}
        \toprule
        \multirow{2}{*}{Methods}  & \multicolumn{3}{c}{Random} & \multicolumn{3}{c}{LLM4TS} & \multicolumn{3}{c}{TEMPO} & \multicolumn{3}{c}{Time-LLM} & \multicolumn{3}{c}{GPT4TS}\\
        \cmidrule(lr){2-4} \cmidrule(lr){5-7} \cmidrule(lr){8-10} \cmidrule(lr){11-13} \cmidrule(lr){14-16}
        & SMAPE & MASE & OWA & SMAPE & MASE & OWA & SMAPE & MASE & OWA & SMAPE & MASE & OWA & SMAPE & MASE & OWA\\ 
        \midrule
\textit{M4-Monthly KernelSynth}  & - & - & - & 14.477 & 1.077 & 1.008 & 13.991 & 1.066 & 0.986 & 13.475 & 1.051 & 0.961 & 14.695 & 1.095 & 1.024 \\
\textit{M4-Quarterly KernelSynth}  & - & - & - & 12.063 & 1.457 & 1.079 & 11.784 & 1.422 & 1.053 & 11.13 & 1.311 & 0.984 & 11.971 & 1.414 & 1.059 \\
\textit{M4-Yearly KernelSynth}  & - & - & - & 16.619 & 3.743 & 0.979 & 16.051 & 3.513 & 0.933 & 13.743 & 3.027 & 0.801 & 17.008 & 3.733 & 0.990 \\
\midrule
\textit{M4-Monthly Ours}  & - & - & - &  13.157 & 0.981 & 0.917 & 12.975 & 0.96 & 0.901 & 13.877 & 1.111 & 1.017 &  13.157 & 0.981 & 0.917  \\
\textit{M4-Quarterly Ours}  & - & - & - &  10.608 & 1.253 & 0.939 & 10.318 & 1.207 & 0.909 & 10.877 & 1.342 & 1.022 & 10.608 & 1.253 & 0.939 \\
\textit{M4-Yearly Ours}  & - & - & - & 15.547 & 3.72& 0.944 & 13.466 & 3.036 & 0.794 & 13.788 & 3.255 & 0.843 & 15.547 & 3.72 & 0.944 \\
\textit{Average}  & - & - & - &  13.104 & 1.985 & 0.933 & 12.253 & 1.734 & 0.868 & 12.847 & 1.903 & 0.961 & 13.104 & 1.985 & 0.933 \\
        \midrule
\textit{M4-Monthly Real} & 22.756 & 1.959 & 1.71 & 12.817 & 0.947 & 0.890 & 12.698 & 0.934 & 0.879 & 13.327 & 1.023 & 0.943 & 12.916 & 0.958 & 0.898\\
\textit{M4-Quarterly Real} & 19.216 & 2.587 & 1.816 & 10.301 & 1.207 & 0.908  & 10.077 & 1.177 & 0.887 & 10.672 & 1.266 & 0.946 & 10.386 & 1.230 & 0.920\\
\textit{M4-Yearly Real} & 37.396 & 8.755 & 2.246 & 13.885 & 3.240 & 0.833 & 13.493 & 3.052 & 0.797 & 13.498 & 3.013 & 0.792 & 14.801 & 3.633 & 0.910\\
\textit{Average} & 26.456 & 4.434 & 1.924 & 12.334 & 1.798 & 0.877 & 12.089 & 1.721 & 0.854 & 12.499 & 1.767 & 0.894 & 12.701 & 1.940 & 0.909\\
    \bottomrule 
    \end{tabular}
    }
    \caption{Time series forecasting results on unseen time series dataset. The forecasting horizons are in [6, 48] and report value is the average. A lower value indicates better performance. Red: the best, Blue: the second best.}
    \label{tab:m4_shorttermforecasting} 
\end{table*}

\begin{table*}[h]
   \centering 
   \begin{tabular}{l|ccc|ccc|ccc}
       \toprule
       \multirow{2}{*}{Frequency} & \multicolumn{3}{c}{LLM4TS} & \multicolumn{3}{c}{TEMPO} & \multicolumn{3}{c}{GPT4TS} \\
       \cmidrule(lr){2-4} \cmidrule(lr){5-7} \cmidrule(lr){8-10}
       & SMAPE & MASE & OWA & SMAPE & MASE & OWA & SMAPE & MASE & OWA \\
       \midrule
       Monthly & \textbf{True} & \textbf{True} & \textbf{True} & \textbf{True} & \textbf{True} & \textbf{True} & \textbf{True} & \textbf{True} & \textbf{True} \\
       Quarterly & \textbf{True} & \textbf{True} & \textbf{True} & \textbf{True} & \textbf{True} & \textbf{True} & \textbf{True} & \textbf{True} & \textbf{True} \\
       Yearly & \textbf{True} & \textbf{False} & \textbf{True} & \textbf{True} & \textbf{True} & \textbf{True} & \textbf{True} & \textbf{False} & \textbf{True} \\
       \bottomrule
   \end{tabular}
   \caption{Statistical significance test results ($p < 0.05$) for differences in SMAPE, MASE and OWA metrics between model trained on our and KernelSynth data based on scores reported in Table~\ref{tab:m4_shorttermforecasting}. Results where our method is better are highlighted in bold.}
   \label{tab:significance-tests} 
\end{table*}


\section{Prototypes Sample Result}
\label{prototypes_domain_sample}

\begin{figure*}
        \centering
        \includegraphics[scale=0.8]{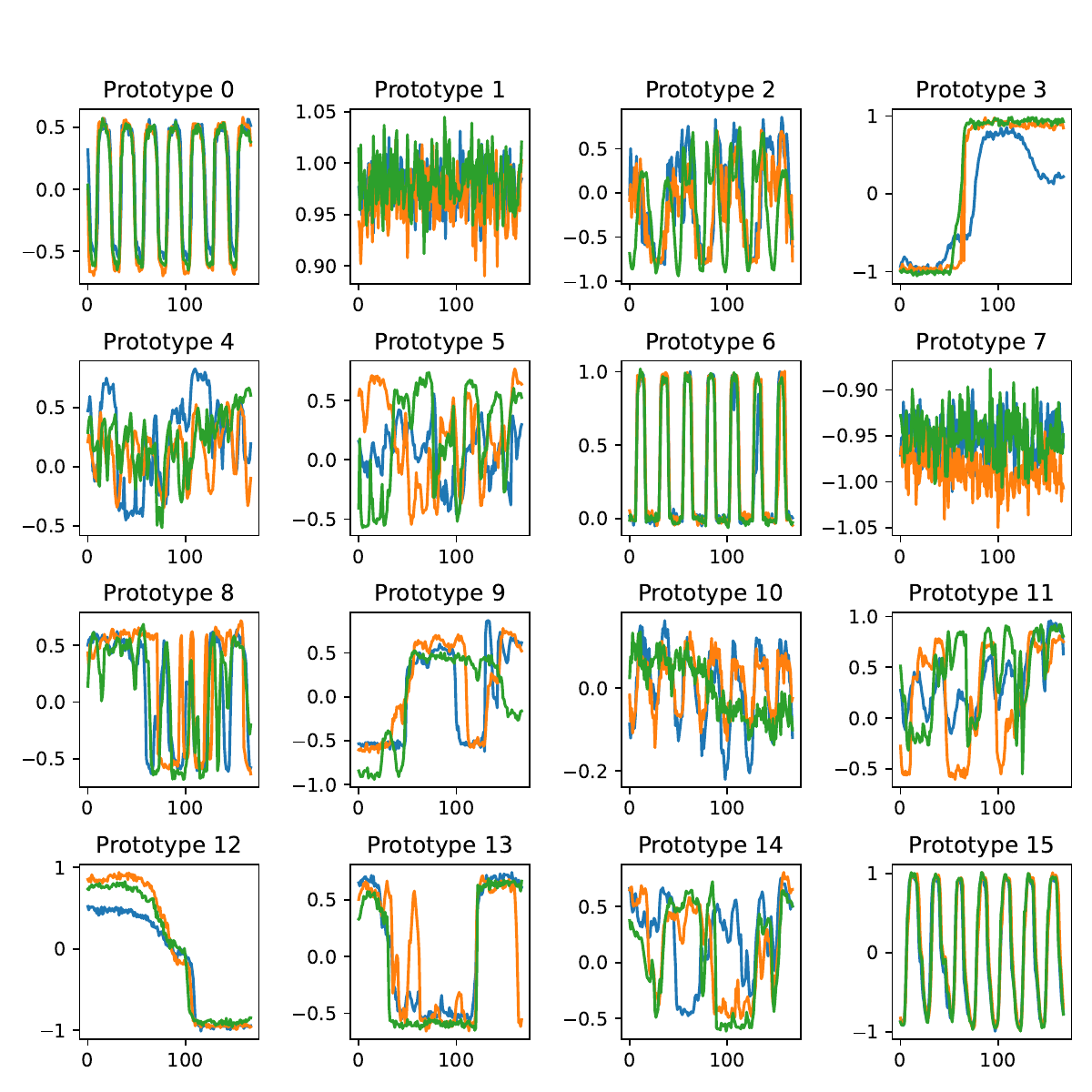}
        \caption{Visualization of semantic prototypes. Each prototype represents a different pattern or characteristic commonly found in time series data. }
        \label{fig:prototype_semantic}
\end{figure*}

\begin{figure}[htbp]
    \centering
    \includegraphics[scale=0.8]{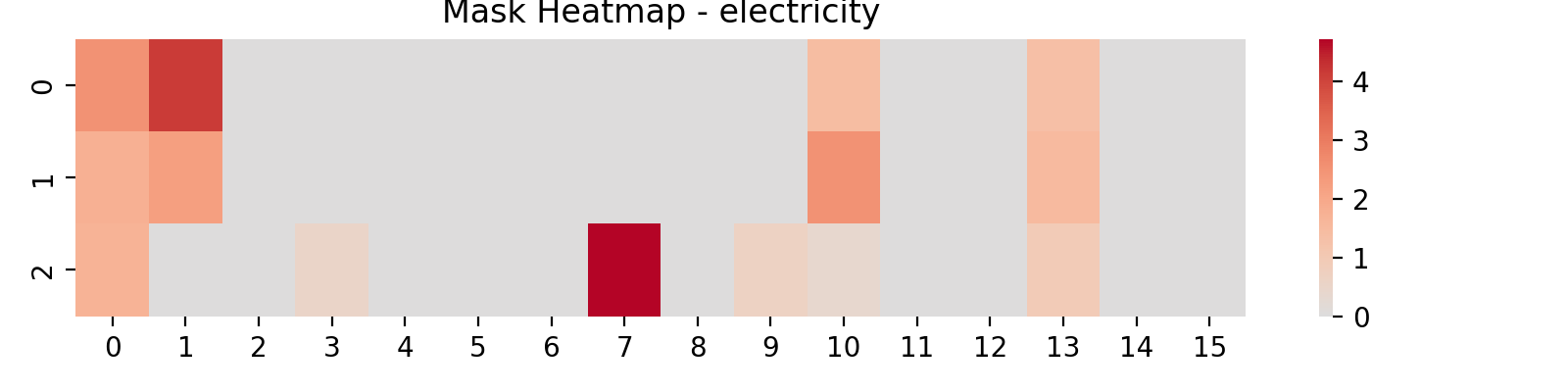}
    \caption{Prototype distribution for the electricity domain. Each heatmap shows prototype indices (x-axis, 0–15) and their frequency or importance (color intensity).}
    \label{fig:prototype_electricity}
\end{figure}

\begin{figure}[htbp]
    \centering
    \includegraphics[scale=0.8]{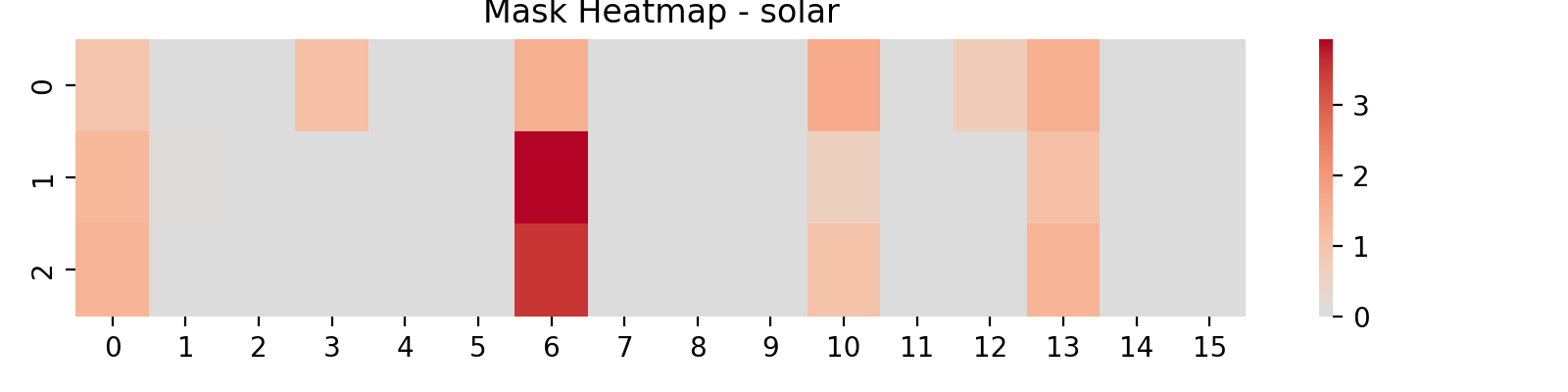}
    \caption{Prototype distribution for the solar domain. Each heatmap shows prototype indices (x-axis, 0–15) and their frequency or importance (color intensity).}
    \label{fig:prototype_solar}
\end{figure}

\begin{figure}[htbp]
    \centering
    \includegraphics[scale=0.8]{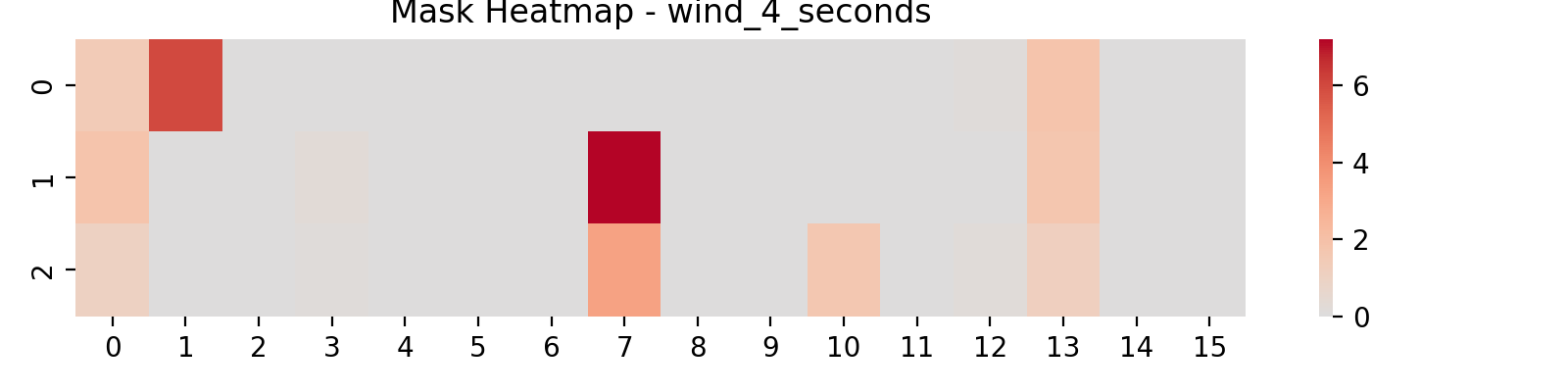}
    \caption{Prototype distribution for the wind (4 seconds) domain. Each heatmap shows prototype indices (x-axis, 0–15) and their frequency or importance (color intensity).}
    \label{fig:prototype_wind}
\end{figure}

\begin{figure}[htbp]
    \centering
    \includegraphics[scale=0.8]{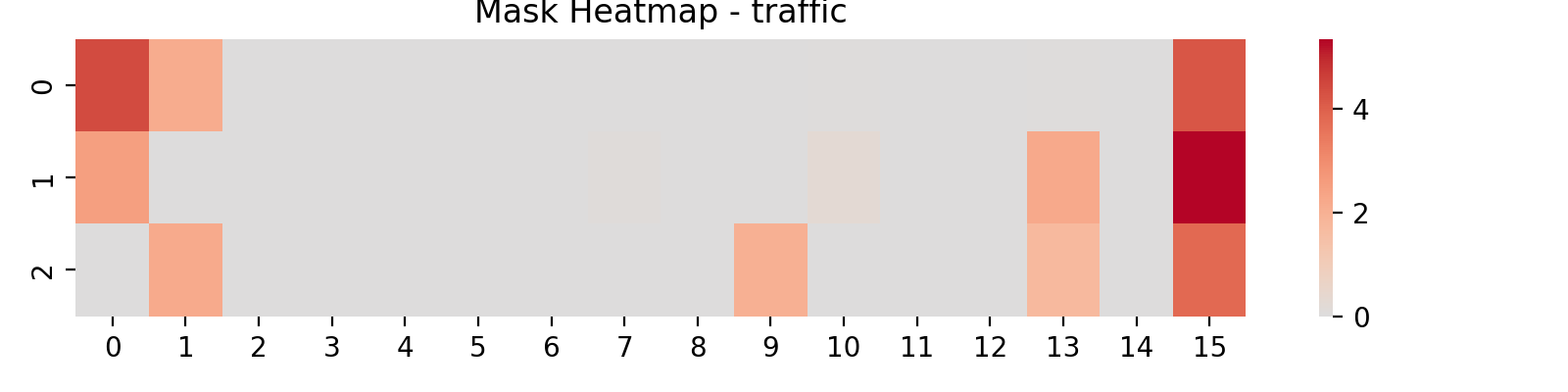}
    \caption{Prototype distribution for the traffic domain. Each heatmap shows prototype indices (x-axis, 0–15) and their frequency or importance (color intensity).}
    \label{fig:prototype_traffic}
\end{figure}

\begin{figure}[htbp]
    \centering
    \includegraphics[scale=0.8]{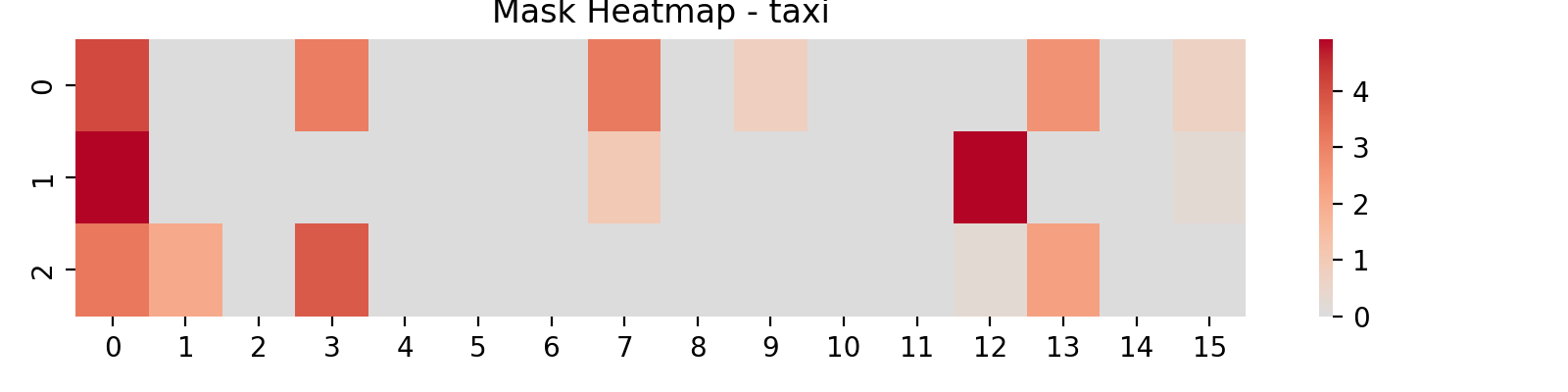}
    \caption{Prototype distribution for the taxi domain. Each heatmap shows prototype indices (x-axis, 0–15) and their frequency or importance (color intensity).}
    \label{fig:prototype_taxi}
\end{figure}

\begin{figure}[htbp]
    \centering
    \includegraphics[scale=0.8]{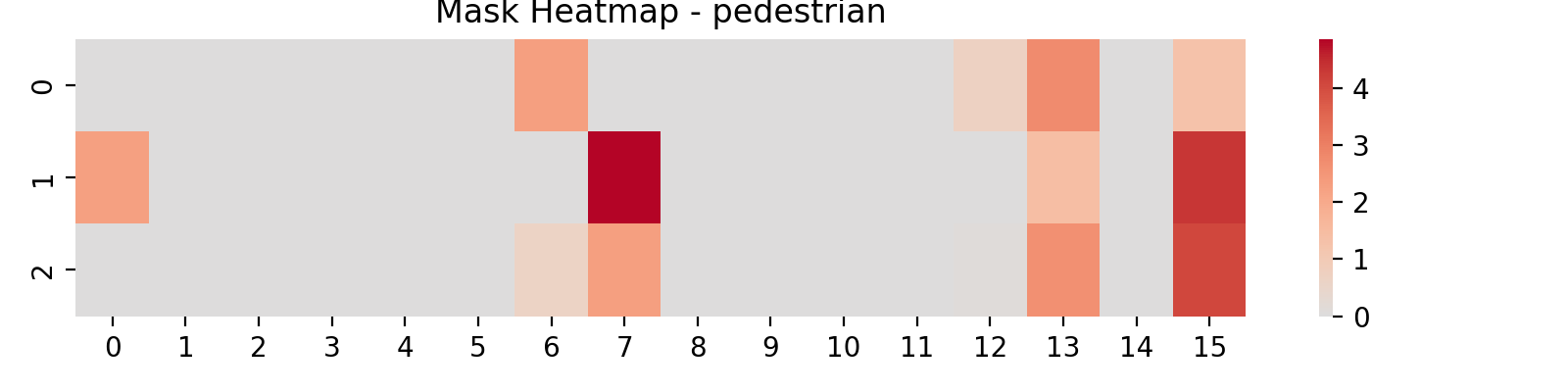}
    \caption{Prototype distribution for the pedestrian domain. Each heatmap shows prototype indices (x-axis, 0–15) and their frequency or importance (color intensity).}
    \label{fig:prototype_pedestrian}
\end{figure}

\begin{figure}[htbp]
    \centering
    \includegraphics[scale=0.8]{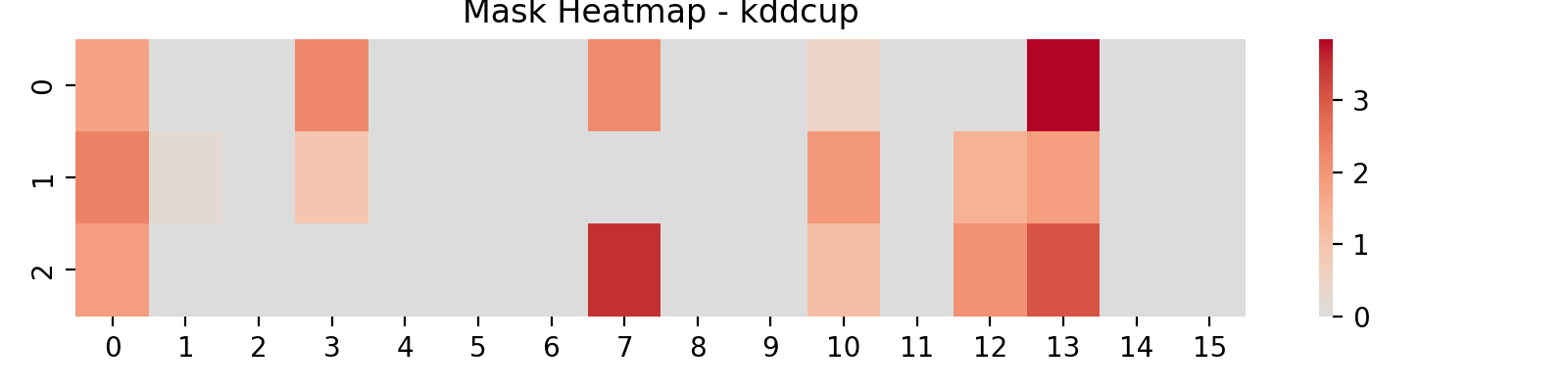}
    \caption{Prototype distribution for the kddcup domain. Each heatmap shows prototype indices (x-axis, 0–15) and their frequency or importance (color intensity).}
    \label{fig:prototype_kddcup}
\end{figure}

\begin{figure}[htbp]
    \centering
    \includegraphics[scale=0.8]{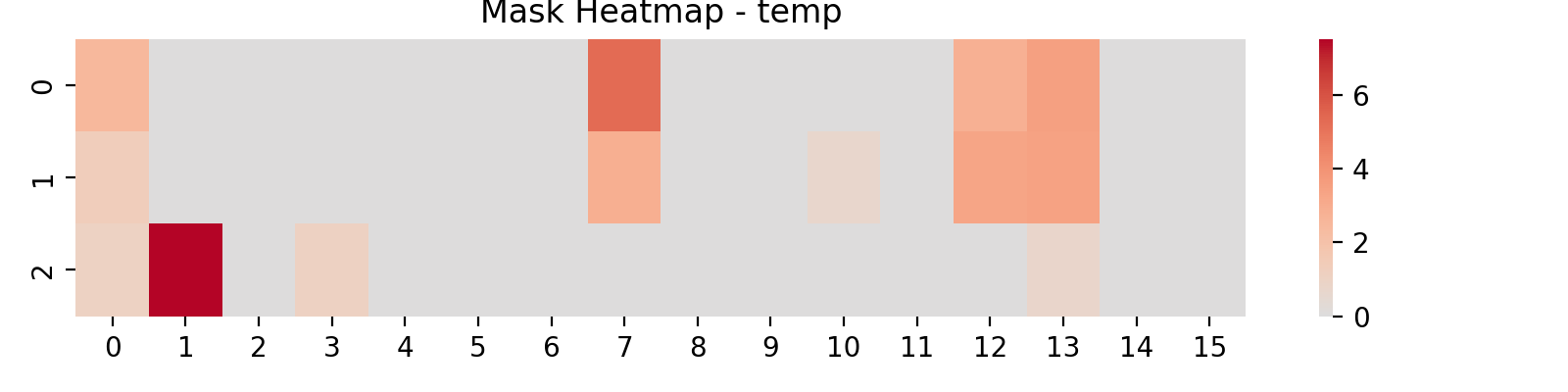}
    \caption{Prototype distribution for the temperature domain. Each heatmap shows prototype indices (x-axis, 0–15) and their frequency or importance (color intensity).}
    \label{fig:prototype_temp}
\end{figure}

\begin{figure}[htbp]
    \centering
    \includegraphics[scale=0.8]{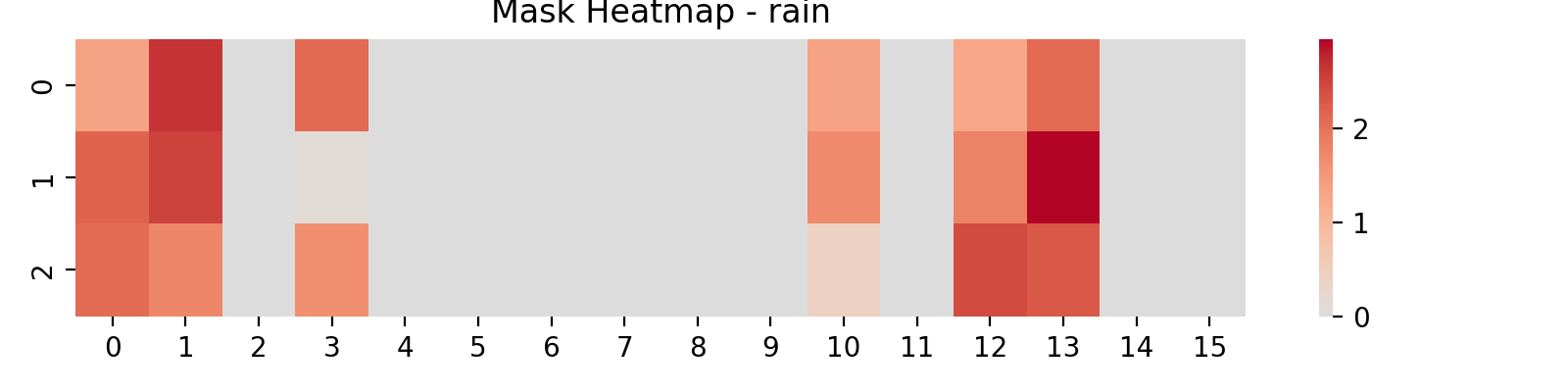}
    \caption{Prototype distribution for the rain domain. Each heatmap shows prototype indices (x-axis, 0–15) and their frequency or importance (color intensity).}
    \label{fig:prototype_rain}
\end{figure}

\begin{figure}[htbp]
    \centering
    \includegraphics[scale=0.8]{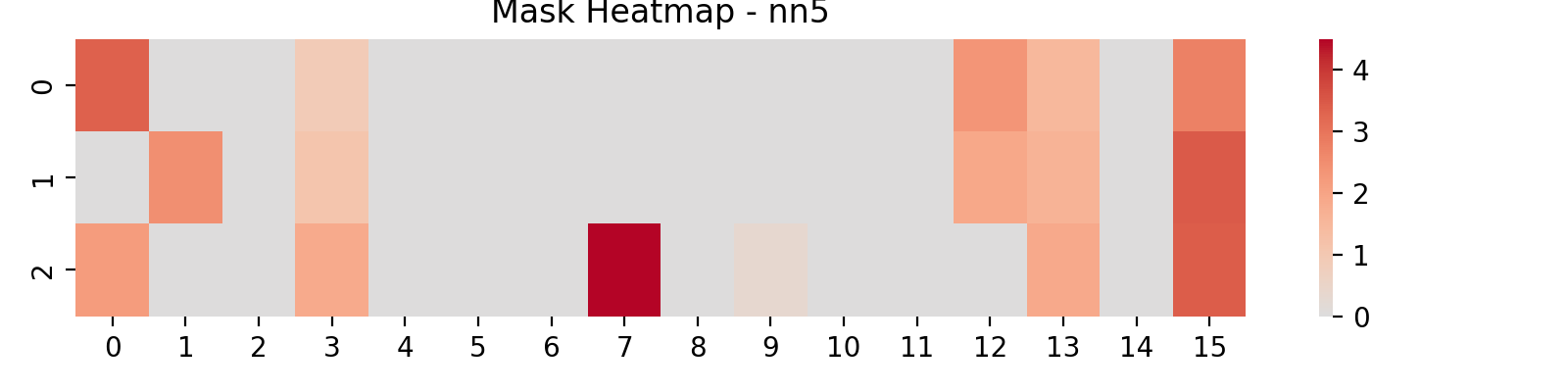}
    \caption{Prototype distribution for the nn5 domain. Each heatmap shows prototype indices (x-axis, 0–15) and their frequency or importance (color intensity).}
    \label{fig:prototype_nn5}
\end{figure}

\begin{figure}[htbp]
    \centering
    \includegraphics[scale=0.8]{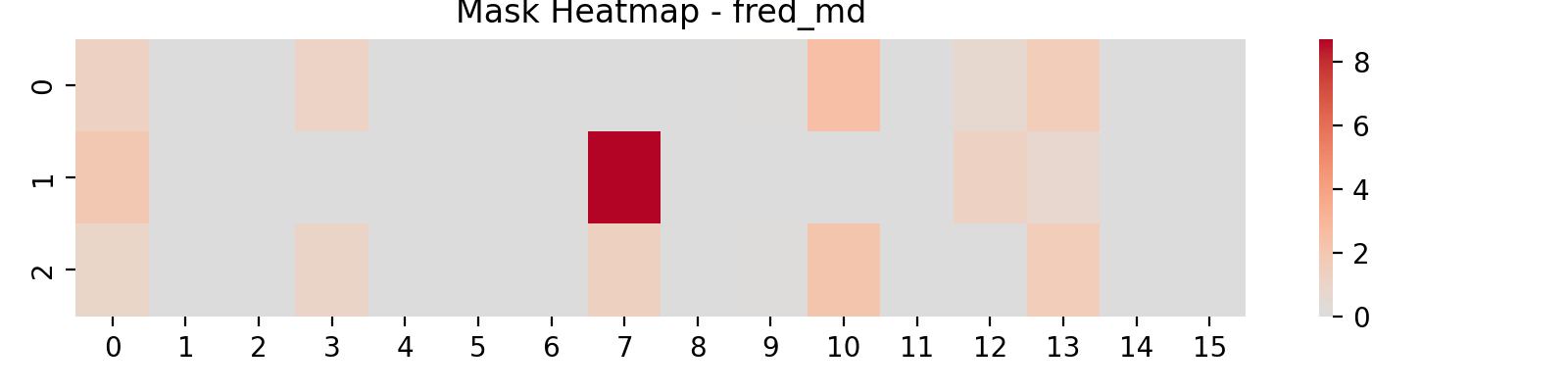}
    \caption{Prototype distribution for the fred\_md domain. Each heatmap shows prototype indices (x-axis, 0–15) and their frequency or importance (color intensity).}
    \label{fig:prototype_fred_md}
\end{figure}

\begin{figure}[htbp]
    \centering
    \includegraphics[scale=0.8]{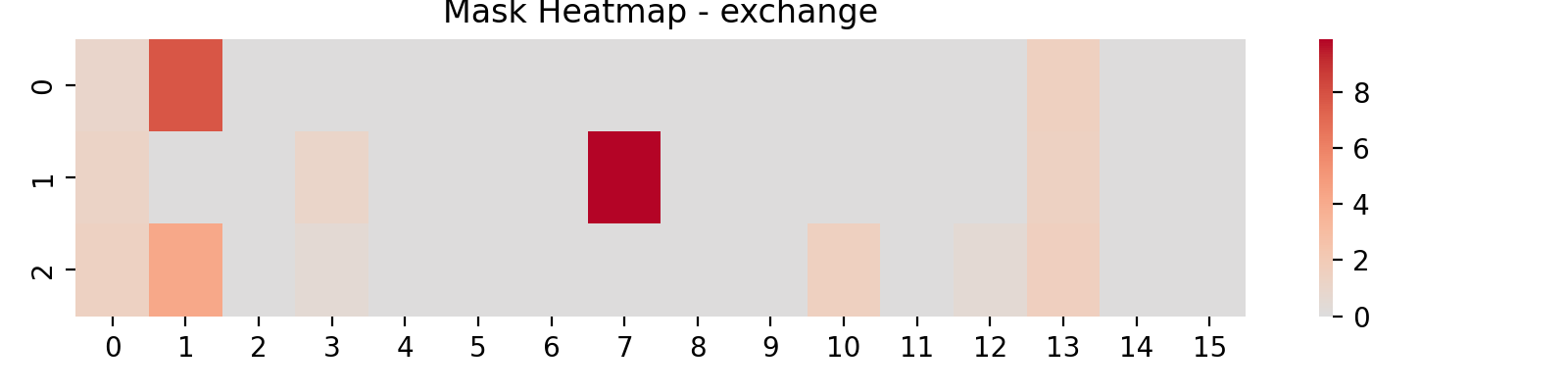}
    \caption{Prototype distribution for the exchange domain. Each heatmap shows prototype indices (x-axis, 0–15) and their frequency or importance (color intensity).}
    \label{fig:prototype_exchange}
\end{figure}

Figure \ref{fig:prototype_semantic} shows 16 semantic prototypes used in our TSG models. Each prototype represents a distinct pattern in time series data, enabling the generation of diverse, domain-specific series. For example, prototypes \{0,6,15\} capture cyclical patterns useful for seasonal trends. Prototypes \{3,12\} represent trend patterns, including gradual changes and sharp transitions. Prototypes \{1,2,7\} show high-frequency fluctuations, representing volatility. By combining these prototypes, the model can generate rich, domain-specific time series data through translating text into time series data with specific semantic concepts. Figure \ref{fig:prototype_electricity} to Figure \ref{fig:prototype_exchange} shows the distribution of prototypes across various domains. Some prototypes, like Prototype 0 in "temperature" and "electricity," are widely relevant, while others, like Prototype 15 in "traffic," are domain-specific. The sparsity of the heatmaps shows that not all prototypes are equally important within a domain. For example, "solar" primarily uses prototypes \{0,6,10,13\}. This demonstrates the flexibility of the prototype-based approach, capturing both general and domain-specific patterns. 

\end{document}